\pdfoutput=1
\documentclass[sigconf,nonacm,9pt]{acmart}
\usepackage[ruled,vlined,noline,linesnumbered]{algorithm2e}
\usepackage{caption}
\usepackage{subcaption}
\usepackage{tabularx}
\usepackage{enumitem}
\usepackage{comment}
\usepackage{balance}
\usepackage{titlesec}
\usepackage{soul}
\usepackage{pifont}
\usepackage{amsmath}
\usepackage[normalem]{ulem}

\newcommand{\cmark}{\ding{51}}%
\newcommand{\xmark}{\ding{55}}%

\AtBeginDocument{%
  }

\setcopyright{acmlicensed}
\acmJournal{TOSN}
\acmYear{2023} \acmVolume{1} \acmNumber{1} \acmArticle{1} \acmMonth{1} \acmPrice{15.00}\acmDOI{10.1145/3623397}

\begin{document}

\title[End-to-end Privacy Protection in Sensing Systems via Personalized Federated Learning]{Blinder: End-to-end Privacy Protection in Sensing Systems via Personalized Federated Learning}

\author{Xin Yang}
\affiliation{%
  \institution{University of Alberta}
  \city{Edmonton}
  \state{AB}
  \country{Canada}}
\email{xin.yang@ualberta.ca}
  \orcid{0000-0002-7861-0007}

\author{Omid Ardakanian}
\affiliation{%
  \institution{University of Alberta}
  \city{Edmonton}
  \state{AB}
  \country{Canada}}
  \email{ardakanian@ualberta.ca}
  \orcid{0000-0002-6711-5502}

\begin{abstract}
This paper proposes a sensor data anonymization model 
that is trained on decentralized data and strikes a desirable trade-off between 
data utility and privacy, even in heterogeneous settings 
where the sensor data have different underlying distributions.
Our anonymization model, dubbed Blinder, is based on a variational autoencoder 
and one or multiple discriminator networks trained in an adversarial fashion.
We use the model-agnostic meta-learning framework to adapt the anonymization model trained 
via federated learning to each user's data distribution. 
We evaluate Blinder under different settings and show that it provides end-to-end privacy protection on two IMU datasets
at the cost of increasing privacy loss by up to $4.00\%$ %
and decreasing data utility by up to $4.24\%$,
compared to the state-of-the-art anonymization model trained on centralized data. We also showcase Blinder's ability to anonymize the radio frequency sensing modality.
Our experiments confirm that Blinder can obscure multiple private attributes at once,
and has sufficiently low power consumption and computational overhead 
for it to be deployed on edge devices and smartphones to perform real-time anonymization of sensor data.
\end{abstract}

\ccsdesc[300]{Computer systems organization~Sensor networks}
\ccsdesc[500]{Security and privacy}

\keywords{Privacy-utility trade-off, %
deep generative models, federated learning} %

\maketitle

\section{Introduction}
Networked, embedded sensors are prevalent in today's urban infrastructure and mobile devices.
The vast amount of data emitted by them and 
the increasing ability to train machine learning models on 
resource-constrained devices have paved the way for numerous applications.
For example, 
timeseries generated by the inertial measurement unit~(IMU)
in a smartphone or wearable can be used to train machine learning models to detect
an individual's activity, health, and vital signs in near real-time~\cite{DeepSense, garcia2018mental, chen2021movi}.
While mobile and edge devices that integrate these sensors 
bring convenience to our lives, they also put our privacy at risk 
as sensor readings can be exposed to an untrusted party with malicious intent~\cite{hajihassani2022, kroger2018unexpected, aloufi2020privacy, plarre2011continuous, malekzadeh2019mobile}.
This adversary can infer an individual's \emph{private attribute}, 
such as gender, race, body size and fitness, 
without their consent, by training a model on data
that was originally gathered for a different purpose, 
e.g., tracking changes in their \emph{public attribute}, such as activity.

\begin{figure}[t]
\centering
\includegraphics[width=\linewidth]{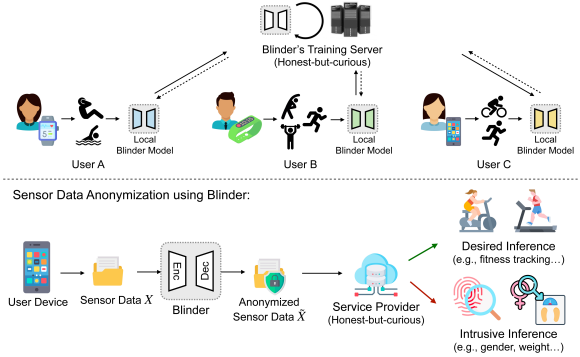}
\caption{Illustration of Blinder's training on decentralized data and its application to anonymize each user's private attribute(s).} %
\label{fig:overview}
\end{figure}

Several methods have been developed to date to support 
privacy-preserving inferences about public attributes of an individual,
while mitigating \emph{attribute inference attacks} that target their private attributes.
They can be broadly classified into data obfuscation and anonymization~\cite{malekzadeh2019mobile,li2021deepobfuscator,hajihassnai2021obscurenet},
differential privacy~\cite{lv2021security,liu2016dependence}, 
secure multi-party computation (SMC)~\cite{li2021privacy, thoma2012secure}, and 
homomorphic encryption~\cite{raisaro2018protecting, chou2020privacy} based methods.
Data obfuscation methods remove or distort patterns 
in the sensor data that contain information about private attributes.
Unlike differential privacy, which is more applicable to a large data repository,
they can be applied to streaming data from one or multiple sensors. 
Moreover, %
they have less computational overhead than SMC and homomorphic encryption, 
making them a natural fit for mobile and edge devices, 
which typically have limited compute power and run on a battery.

Recent advances in deep generative models have made possible the development 
of sophisticated anonymization models that are more suitable for embedded sensor systems.
For instance, Anonymization AutoEncoder~\cite{malekzadeh2019mobile}, 
Olympus~\cite{raval2019olympus}, 
and ObscureNet~\cite{hajihassnai2021obscurenet} 
use autoencoders and adversarial training to learn a representation of sensor data 
that can be used to artificially generate anonymized sensor data.
Since public and private attributes are usually entangled in the latent space 
(e.g., %
slow movement can indicate both activity and age group),
modifying the learned latent representation to obscure an individual's private attribute
often lowers its utility for the desired inference task, 
resulting in a trade-off between privacy and data utility for each individual.
Training an anonymization model that yields the best privacy-utility trade-off 
is a difficult task, especially for a diverse set of users. 
On the one hand, an anonymization model trained on a small number of samples 
pertaining to a single user may not generalize well to new samples due to overfitting.
Moreover, public or private attribute classes are more likely 
to have an unequal representation in the training dataset when it includes samples from a single user only 
(e.g., an individual who bikes regularly may run only occasionally).
This diminishes the ability of the anonymization model to generalize to new samples.
On the other hand, training an anonymization model on a large dataset 
to tackle non-i.i.d. data and attribute imbalance issues
would require many users to send their raw sensor data along with their attributes to a central server.
This server might be a semi-honest or honest-but-curious (HBC) adversary~\cite{goldreich2009foundations},
which performs computation correctly but ``on the side'' uses the obtained data to their advantage
(e.g., to monetize the users' private attribute\footnote{Many companies presently collect and monetize
their users' personal data~\cite{Kirkpatrick22}, from web browsing and shopping habits 
to lifestyle and health conditions inferred from their sensor data, which is the focus of this paper. 
They can be viewed as an HBC adversary.}). 
Existing anonymization models either do not generalize well to a large heterogeneous population
or are trained on large centralized data failing to protect the privacy of users 
who contributed their sensor data.
This calls for a data anonymization technique that protects the privacy of users,
both during the model training phase and after it is deployed to anonymize sensor data.

Federated learning allows a set of users to cooperatively train a model,
in this case a powerful data anonymization model,
without exchanging their local data and private attributes.
By updating parameters of the local model on the respective device,
it ensures that the user's privacy is protected while the model is being trained.
However, the standard Federated Averaging algorithm (FedAvg)~\cite{mcmahan2017communication} 
cannot be adopted in heterogeneous settings, 
where users' data distributions are not identical, 
i.e., local data fail to represent the overall distribution~\cite{li2020federated, karimireddy2020scaffold}.
This is because local gradient estimates are biased 
when the users' sensor data have different distributions,
and performing aggregation on the biased estimates 
could degrade the performance of the anonymization model.

In this paper, we propose \textit{Blinder}, a novel data anonymization model based on 
a variational autoencoder and %
discriminator network(s).
By using adversarial training, and appending public and private attributes to the learned latent representation,
we deliberately structure the latent space of the autoencoder 
to generate the anonymized version of sensor data.
We show that the anonymized sensor data contains nearly the same amount of information about the public attribute, 
yet much less information about the private attribute
than the original sensor data (see Section~\ref{subsec:external_validation}).

Blinder is trained using a \emph{personalized federated learning} algorithm 
to address data heterogeneity and class imbalance issues.
Specifically, we apply \emph{meta-learning} to federated learning 
to enhance generalizability and enable personalization 
of the anonymization model~\cite{fallah2020personalized, collins2021exploiting}.
By offloading model training to the client side, 
users that participate in the training phase are not required 
to share their raw sensor data and attributes with a central server.
The high adaptability of the meta-learned anonymization model makes it effective 
even for new users who did not previously take part in the training phase.
They can fetch parameters of the fully trained Blinder
from the server and perform 
adaptation on a small portion of their local data %
before using this personalized model to anonymize their sensor data. %
We summarize our main contributions below:
\vspace{-\topsep}
\begin{itemize}
    
    \item 
        We develop a novel sensor data anonymization model 
        based on a variational autoencoder and one or multiple discriminator networks. 
        Being trained via a meta-learning-based federated learning algorithm, 
        Blinder is the first anonymization model that offers end-to-end privacy protection 
        while achieving high data utility in heterogeneous settings.
        We tackle the class imbalance issue that we face during model training 
        by adopting re-sampling strategies to balance the distribution of public attributes.
        To balance the distribution of private attributes, 
        even in the extreme case where there are missing private attribute classes in a user's data,
        we synthesize \emph{shadow samples} by taking advantage of Blinder's decoder.
        These techniques pave the way for training Blinder using a personalized federated learning framework, improving its ability to learn from non-i.i.d. data and 
        generalize to unseen users that might have different data distributions.

    \item 
    We evaluate Blinder on two IMU datasets and one Wi-Fi sensing dataset, and compare it with four baselines. 
    Our results show that Blinder provides strong privacy protection 
    in the entire data consumption life cycle. 
    We show that it can effectively reduce the accuracy of intrusive inferences 
    (a measure of \emph{privacy loss}) about one or multiple private attributes
    without a substantial reduction in the accuracy of desired inferences 
    (a measure of \emph{data utility}).

    \item We deploy Blinder
    on 3~Android smartphones and 1~NVIDIA Jetson Nano 
    (representative of an IoT edge device)
    and conduct real-world experiments to corroborate 
    that Blinder has low power consumption and can perform real-time anonymization of data 
    emitted by the embedded sensors.\footnote{Our implementation of Blinder and the mobile application is available on GitHub: \url{https://github.com/sustainable-computing/blinder}.} %
    
\end{itemize}

\section{Related Work}
\subsection{Privacy-Preserving Data Analytics}
Significant progress has been made in recent years towards developing privacy-preserving techniques 
for mobile and IoT devices that are equipped with various sensors.
Differential privacy (DP)~\cite{dwork2006our, abadi2016deep} provides 
mathematically verifiable privacy guarantees by adding perturbations to the data. 
However, without factorizing the private attribute(s) and selectively introducing perturbations,
applying DP to sensor data could significantly deteriorate its utility.
Techniques based on SMC~\cite{goldreich1998secure, so2021codedprivateml}
and homomorphic encryption utilize secret sharing algorithms %
and cryptographic methods to provide information-theoretic security
such that the participating parties can perform computation 
based on a portion of encrypted data, without being able to reconstruct the original data.
Despite these advantages, SMC and homomorphic encryption have large computational overhead 
and can quickly drain the battery life of mobile devices~\cite{gilad2016cryptonets, yang2021secure}.

Recent progress in deep and adversarial learning~\cite{zhao2020trade,hamm2017minimax} has brought a new perspective to data anonymization.
Deep neural networks~\cite{lecun2015deep} can be embedded in the sensor data processing pipeline
to hide or remove patterns in timeseries that can be correlated to 
private attributes with moderate computational overhead. 
Liu \textit{et al.}~\cite{liu2019privacy} propose Privacy Adversarial Network~(PAN), 
an encoder trained in an adversarial fashion to produce feature representations 
that do not contain sensitive user information.
Li \textit{et al.} propose DeepObfuscator~\cite{li2021deepobfuscator} 
which jointly trains a convolutional neural network (CNN) comprising a feature extractor 
and the desired task classifier, together with two adversarial network components 
designed to reconstruct data and predict private attributes. 
Rather than extracting anonymized feature representations, 
Malekzadeh \textit{et al.}~\cite{malekzadeh2019mobile} and Hajihassnai \textit{et al.}~\cite{hajihassnai2021obscurenet}
propose the use of an autoencoder-based architecture %
to obscure private attributes in the latent space of an autoencoder.
In similar work, Olympus~\cite{raval2019olympus} jointly optimizes the utility loss and inference accuracy 
on top of an autoencoder to achieve utility-aware data anonymization.
Bertran \textit{et al.}~\cite{bertran2019adversarially} use 
adversarial neural networks to match the distributions of the public and private attributes, 
and perform data obfuscation without changing the format of input data.
We categorize these models as \emph{format-preserving data anonymization} 
for they retain the original data format, 
allowing legacy applications to seamlessly consume the anonymized data. %
Nevertheless, they are only useful when the training dataset is publicly available 
or users that participate in the anonymization model training are willing to give up on their privacy.
In contrast, Blinder protects users' privacy even when the 
anonymization model is being trained or updated.

Distributed privacy-preserving techniques eliminate the need for 
sharing (potentially private) user data with a service provider 
by taking advantage of local computation.
There have been substantial efforts in recent years~\cite{truex2019hybrid, xu2019hybridalpha} 
to utilize federated learning (FL) to avoid sharing private user data with untrusted servers during model training. 
However, these papers assume that application developers 
will redesign the application to fit their federated learning framework. 
Mo~\textit{et~al.}~\cite{mo2021ppfl} study the privacy risks in FL frameworks 
and propose deploying FL onto Trusted Execution Environments (TEE) 
to preserve privacy and defend against data reconstruction attacks with a small system overhead.
While TEEs can become prevalent in the future,
we take a different approach and 
develop a privacy-preserving framework that does not require additional hardware.
TIPRDC~\cite{li2020tiprdc} is a privacy-preserving data crowdsourcing framework 
enabled by training a feature extractor through an adversarial game to conceal private attributes. 
It maintains data utility by maximizing the mutual information 
between raw data and the combination of the private attribute and extracted feature. 
Yet, TIPRDC extracts privacy-preserving feature representations, 
hence developers must update their application for it to be compatible with the anonymized features.
It differs from our data anonymization model which generates an anonymized version of 
data that has the same dimensions as its input (both are in the same space), 
allowing existing applications to readily use the anonymized data without any modification.
Table~\ref{tab:compare} compares Blinder and the previous work 
on privacy protection that uses machine learning techniques.

\subsection{Federated Learning and Personalization}
Federated learning~\cite{mcmahan2017communication} is a standard framework 
for training machine learning models on decentralized data. 
While the FedAvg algorithm can achieve good performance 
when users' local datasets have the same underlying distribution,
it may not work when they contain imbalanced classes or their features are non-i.i.d. 
    Data augmentation-based techniques are especially effective in addressing the class imbalance issue~\cite{chawla2002smote,kuhn2013remedies,japkowicz2002class,drummond2003c4} 
    through oversampling the minority class or downsampling the majority class. 
    However, balancing the class distribution across all local datasets
    might require the server to know the class distribution 
    in each local dataset a priori.
    Some previous work proposes creating a new layer,
    in addition to the aggregation server and users, to balance the class distribution more effectively.
    For example, Wang~\textit{et~al.}~\cite{wang2021addressing} note that the magnitude of 
    gradient updates correlates with the number of samples in each class, 
    so they introduce a monitor that infers 
    the overall class distribution across all local datasets in each round of FL. 
    Upon detection of an imbalance aggregation, a ratio loss is computed 
    to balance the contribution of each class in the probability prediction.
    Duan~\textit{et~al.}~\cite{duan2020self} propose Astraea, a self-balancing FL framework that 
    combines data augmentation and a set of mediators between the clients and aggregation server. 
    The mediators reschedule the training process based on the clients' data distribution, 
    allowing a group of users holding a collection of data with partial equilibrium 
    to participate in training sequentially.
    These approaches require redesigning the existing FL framework to incorporate the intermediate layer.
    FedSens~\cite{zhang2021fedsens} eliminates the intermediate layer by training 
    a reinforcement learning agent on the edge to determine whether the edge device 
    should perform a local model update and send it to the server. 
    Meanwhile, the aggregation server adaptively controls the global update frequency 
    based on the received local updates.
    However, none of the above papers discusses the case where some classes can be completely missing.
    BalanceFL~\cite{shuai2022balancefl} considers the missing class issue and 
    proposes a distillation loss that allows the local model 
    to inherit the knowledge of missing classes from the global model. 
    This work ensures the local updates are ``pseudo-uniform'' by applying balanced sampling, 
    data augmentation, and regularization terms. 
    Thus, the global model aggregated on nearly balanced local updates could achieve 
    the ideal performance even using the vanilla FedAvg algorithm.

The non-i.i.d. feature issue is more challenging and may concurrently occur with the class imbalance issue.
To address this problem, several papers have investigated model personalization in federated learning~\cite{fallah2020personalized,jiang2019improving,khodak2019adaptive, arivazhagan2019federated,t2020personalized,deng2020adaptive,collins2021exploiting,li2021hermes}. 
For instance, FedRep~\cite{collins2021exploiting} 
learns a shared neural network model that extracts common feature representations using FL, 
then allows users to add a few layers on top of the shared network 
to enable personalized learning on heterogeneous data.
Dinh~\textit{et~al.}~\cite{t2020personalized} propose pFedMe, 
which uses the Moreau envelope as a regularization term and allows users to train a personalized model 
for their local data distribution in parallel to the global model.
Similarly, Deng~\textit{et~al.}~\cite{deng2020adaptive} enable model personalization 
by mixing the parameters of the global model and a user's local model through a weighted summation, 
where the weights are updated in each communication round.
Both approaches show acceptable performance on heterogeneous datasets 
for users who contributed to the model training, 
but it is unclear whether the learned global model can efficiently adapt to data from
unseen users who might have a different feature distribution.
To our knowledge, none of the above papers uses federated learning 
to train an anonymization model that provides a reasonable trade-off 
between privacy and data utility, while accounting for data heterogeneity.

Meta-learning~\cite{finn2017model, nichol2018first} is a powerful ``learning to learn'' approach 
that can be used to train a highly adaptive model by learning from multiple tasks, 
each with limited data samples.
Applying meta-learning to federated learning makes it possible to forgo data sharing, 
while maintaining fast convergence on small data batches with enhanced adaptation capability (see~\cite{chen2018federated, aramoon2021meta, fallah2020personalized, al2021data}).

With growing applications of generative adversarial networks (GAN), some efforts have been made to train a GAN on decentralized data using federated learning. Fan \textit{et~al.}~\cite{fan2020federated} study the federated training of GAN considering different data skewness and server-client synchronization settings. The authors show that when the data distribution is highly skewed, the performance of FedAvg is unsatisfactory due to weight divergence. Similar observations have been made in~\cite{li2022ifl}, where updates of local generators are aggregated using maximum mean discrepancy (MMD) to improve the performance of FL-GAN~\cite{hardy2019md} in the presence of non-i.i.d. data. GANs have also been employed to synthesize local data samples, thereby enabling personalized federated learning~\cite{cao2022perfed}. However, in that work, the GAN itself is trained on the user device rather than on decentralized data using personalized federated learning. Although the above-mentioned papers provide useful insight into training a generative model on non-i.i.d. data in the FL setting, they neither investigate training a generative model using a personalized federated learning algorithm, nor do they consider the most extreme case of non-i.i.d. data where some classes are completely missing.

\noindent\textbf{Novelty of this paper:} 
Despite recent efforts to enable adaptation in the federated learning framework, 
no related work applies a personalized federated learning algorithm 
to train a complex generative model for data anonymization on decentralized user data.
This problem presents unique challenges that are not encountered 
when training a general machine learning model via federated learning in heterogeneous settings. 
Specifically, in addition to having different feature distributions, users typically have heavily imbalanced public attributes, and a fixed set of private attributes (i.e., the issues of missing classes described in Section~\ref{sec:rebalance}).

\begin{table}[tb]
    \centering
    \resizebox{\columnwidth}{!}{
    \begin{tabular}{| l | c |c |c |c| } 
    \hline
     \textbf{Related Work} & End-to-end Privacy & Compatible w/ Legacy Apps & Require Hardware \\
     \hline
     Privacy-Preserving Feature Extraction~\cite{li2021deepobfuscator,liu2019privacy, li2020tiprdc,hamm2017minimax,zhao2020trade} & \xmark & \xmark & \xmark \\
     \hline
     Format-Preserving Data Anonymization~\cite{raval2019olympus,hajihassnai2021obscurenet,malekzadeh2019mobile, bertran2019adversarially} 
     & \xmark  & \cmark & \xmark  \\
     \hline
     Privacy-Preserving Federated Learning~\cite{truex2019hybrid,xu2019hybridalpha} 
     & \cmark & \xmark & \xmark \\
     \hline
     Privacy-Preserving Federated Learning in TEE~\cite{mo2021ppfl} 
     & \cmark & \xmark & \cmark \\
     \hline
     Blinder (Our Approach) 
     & \cmark & \cmark & \xmark \\
     \hline
    \end{tabular}%
    }
    \vspace{2mm}
    \caption{Comparison between Blinder and prior work on machine learning-based privacy protection.
    }\label{tab:compare}%
    \vspace{-9mm}
\end{table}

\section{Problem Definition}
\label{sec:problem_def}
We represent the dataset that pertains to user $\mathcal{U}_i$ and 
is stored on their device as $\mathcal{D}_i =\{(X, \bar{Y}, Y)\}$, 
where $X$ is one segment of their sensor readings, %
$\bar{Y}$ is the corresponding public attribute,
and $Y$ is the corresponding private attribute that the user would like to conceal.
Data from $N$ users constitute the dataset $\mathcal{D}=\{ \mathcal{D}_1, ..., \mathcal{D}_N\}$. 
For example, in the Human Activity Recognition~(HAR) task, 
the data points generated by $\mathcal{U}_i$'s IMU sensor are denoted $X$, 
the user's activity is denoted $\bar{Y}$,
and their private attribute, such as weight or gender, is denoted $Y$.
The user's private attribute might be immutable, e.g., their race or weight,\footnote{Although 
an individual's weight can change over time, 
the timescale of its changes is much slower than the rate at which sensor data is generated, 
so it can be deemed immutable for the purpose of this study.}
or vary over time, e.g., their geographical location.
In this work, we assume that the user's public and private attributes are categorical variables, 
taking values from a finite set,
and that there is at least one user-defined private attribute that must be protected.

Our goal is to anonymize sensor data in a mobile or edge device
such that the leakage of private information is minimized while data utility is maintained.
The end-to-end privacy protection is achieved in two parts.
The first part is a data anonymization model that effectively obscures the private attribute(s)
so that it cannot be inferred by an adversary that has access to this data.
This model takes as input the raw sensor data $X$, 
obscures the private attribute(s), and produces anonymized data $\widetilde{X}$.
Ideally, unwanted inferences of the private attribute $Y$ from $\widetilde{X}$ 
should attain the same accuracy as random guessing. 
Meanwhile, $\widetilde{X}$ should maintain its utility 
for the task of inferring the public attribute, $\bar{Y}$.
The second part is to eliminate the need for sharing users' raw sensor data $\mathcal{D}$
for training the anonymization model.
We propose to distribute and offload the anonymization model training to users' local devices, 
then perform model aggregation to build an adaptable global anonymization model. 

\paragraph{Adversary Model}
{We consider a practical semi-trusted setting, illustrated in Figure~\ref{fig:overview}, 
in which the service provider or application that performs desired inferences on the user's anonymized sensor data $\widetilde{X}$, and the central server involved in training Blinder
are both an HBC adversary (aka passive adversary defined in~\cite{Goldreich87, goldreich2009foundations}).
This adversary uses a parametric model %
to correctly perform operations that are required 
to complete the assigned task,
e.g., making the desired inference about the user's public attribute given $\widetilde{X}$.
However, the adversary has an incentive to take a peek at the shared data %
and use another machine learning model, %
i.e., the intrusive inference model, 
to identify the user's private attribute $Y$ given $\widetilde{X}$.
We assume both models are pre-trained on a batch of raw sensor data, 
which can be gathered by the adversary.
Additionally, the adversary may have access to the true private attribute 
that corresponds to some data,
enabling it to perform a powerful \emph{re-identification attack}~\cite{hajihassnai2021obscurenet}.
We expand on this in Section~\ref{sec:anon-process}.

}

\section{Blinder's Architecture}
\label{subsec:deep_generative_model_for_data_anonymization}

We present Blinder's architecture and describe the role of each neural network in this architecture.
The autoencoder is a generative model known for its superb ability to extract latent representations. 
The variational autoencoder~(VAE)~\cite{kingma2014semi} introduces regularization of the latent space 
to ensure continuity and completeness, the properties that are essential for generating meaningful new data.
Our motivation for building Blinder on top of the VAE architecture is that 
data anonymization can be viewed as a data generation task 
subject to a constraint that the newly constructed data must contain sufficient information 
about the public attribute and very little information about the private attribute(s).
To enable this, we condition the decoder on public and private attributes, %
and extend the VAE architecture using an auxiliary neural network (i.e., a discriminator) 
per private attribute to allow for adversarial training, 
which is essential for avoiding the information leak 
about the private attribute(s) through latent variables.

For simplicity, we discuss obscuring a single private attribute in this section. 
It is a straightforward extension to simultaneously anonymize multiple private attributes 
by adding more discriminator networks and conditioning the decoder on all of them
(see the discussion in Section~\ref{sec:multiple}). 

\subsection{A Generative Model for Anonymization}
\begin{figure}[t]
\centering
\includegraphics[width=0.8\linewidth]{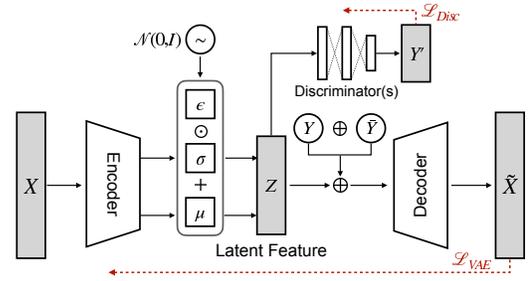}
\caption{Architecture of Blinder's anonymization model.} %
\label{fig:architecture}
\end{figure}

Blinder consists of three neural networks: 
probabilistic encoder, decoder, and discriminator networks as depicted in Figure~\ref{fig:architecture}. 
For anonymizing timeseries sensor data, 
we construct the encoder and decoder by stacking multiple fully connected layers, 
which are parameterized by $\theta$ and $\phi$, respectively.
The encoder network takes as input segmented timeseries data $X$ from one or multiple sensors,
and learns a multivariate Gaussian distribution with mean $\mu$ and log-covariance $\sigma$
over latent representations.
The latent representation of $X$, denoted by $Z$, is randomly sampled from this distribution.
A random variable $\epsilon$ sampled from the standard Gaussian distribution 
is incorporated in the VAE to reparameterize the latent features. 
This step introduces randomness into the data reconstructed by the decoder 
and makes it possible to learn $\mu$ and $\sigma$ through backpropagation~\cite{kingma2013auto}.
The latent representation is derived from $X$ and $\epsilon$:
\begin{equation}
\label{eq:reparam}
    Z = \mu{(X;\theta)} + \epsilon \cdot e^{\frac{\sigma{(X;\theta)}}{2}},\quad \epsilon\sim{\mathcal{N}(0,I)}.
\end{equation}
The reparameterized latent features are then fed to both the decoder and discriminator.
We first describe the decoder which aims to reconstruct the input data from its latent representation $Z$;
we denote the reconstructed data segment by $\widetilde{X}$.
In Blinder, instead of having the decoder generate $\widetilde{X}$ from $Z$, 
we send $Z \oplus Y \oplus \bar{Y}$ to the decoder, where $\oplus$ is the concatenation operation, $Y$ %
and $\bar{Y}$ are private and public attributes associated with $X$, respectively.
We pass $Y$ as a separate feature to the decoder to make the distribution of the private attribute
in the reconstructed data $\widetilde{X}$ depend on the given $Y$.
Thus, in the anonymization stage, we can manipulate $Y$ to 
reconstruct data samples that contain information about a different private attribute, 
concealing the true private attribute of the user.

In addition to the private attribute $Y$, 
Blinder relies on the public attributes to condition the latent features, 
such that a single anonymization model can learn a reconstruction strategy 
that works best for every public attribute label.
Intuitively, different patterns in timeseries data may contain information about 
the private attribute depending on the public attribute label.
For example, suppose our goal is to anonymize gender (private attribute) 
while maintaining data utility for human activity (public attribute) recognition.
The most significant physiological and behavioral differences between the two genders 
can be the whole-body movements when users walk, 
but only upper-limb movements when they sit down.

Finally, we describe the role of the discriminator module.
Observe that since the VAE conditions the latent representation $Z$ on the private attribute $Y$, 
it is possible to change the original $Y$ to a fictitious $\widetilde{Y}$ 
to generate $\widetilde{X}$ that contains information about a different private attribute. %
However, there are no guarantees that the original private attribute
cannot be inferred from the latent representation itself.
We employ an adversarial network in the model training 
to scrub information about the private attributes from the latent representation $Z$.
Specifically, a multi-layer perceptron (MLP), parameterized by $\eta$, is included in Blinder 
to simulate an adversary who wishes to infer the private attribute from the latent representation.

\subsection{Loss Function}
\label{subsec:loss_func}
We introduce different terms in the loss function that are optimized %
to train Blinder.
The first term is the standard VAE loss. 
Instead of maximizing the marginal likelihood which is intractable, 
the VAE is trained to maximize a lower bound on the marginal log likelihood, 
called the evidence lower bound (ELBO)~\cite{kingma2013auto}, 
and is given by $\mathbb{E}_{Q_{(Z|X)}} [{\log P(X,Z)}-{\log Q(Z|X)}]$.
ELBO has two terms: the reconstruction loss and a regularization term. 
The reconstruction loss uses the expected log likelihood 
to measure the similarity between the raw data $X$ and the reconstructed data $\widetilde{X}$.
The regularization term enhances the model generalizability and punishes overfitting 
by encouraging the representations to distribute across the latent space.
We use L2 loss as the reconstruction loss and Kullback–Leibler~(KL) divergence for regularization.
Thus, the VAE's loss function can be written as:
\begin{equation}
 \mathcal{L}_{VAE}(f_{\langle \theta,\phi \rangle})\!\!=\!\!-\mathbb{E}_{Z \sim Q_\theta} \left[ \log{P_\phi}(X|Z,Y,\bar{Y})\right] + D_{KL} (Q_\theta(Z|X) || P_\phi(Z)),
\end{equation}
where $D_{KL}$ is the KL divergence term that measures the distance between the prior latent distribution $P(Z)\sim{\mathcal{N}(0, I)}$ and the posterior distribution $Q_\theta(Z|X)$, 
$f_{\langle \theta,\phi \rangle}$ is the parameterized mapping function for the encoder and decoder,
i.e. $\widetilde{X}=f_{\langle \theta,\phi \rangle}(X)$.
The VAE loss is the total loss of the encoder and decoder networks 
and can be backpropagated and optimized via gradient descent.

The next term in the loss function 
quantifies the performance of the discriminator.
It calculates the cross-entropy between the inferred private attribute $Y'$ and 
the actual private attribute $Y$ associated with the input data segment.
We express the discriminator's loss as:
\begin{equation}
    \mathcal{L}_{Disc}(f_\eta)= - \sum_{Y} Y \log P_\eta(Y') = - \sum_{Y}  Y \log \mathbb{E}_{Z \sim Q_\theta}\big[P_\eta(Y'|Z)\big],
\end{equation}
where %
$f_\eta$ is the mapping function of the discriminator parameterized by $\eta$. 
The adversarial training process can be viewed as a minimax game, 
where the probabilistic encoder in the VAE aims to extract a latent representation from the input data segment
such that the decoder can reconstruct the original data, 
but the discriminator cannot infer the private attribute from this representation.
To encourage the encoder to learn latent representations that are not correlated 
with the private attribute and cannot be easily linked back to it,
we add the negative loss of the discriminator to the VAE loss and formulate the total loss function 
that will be minimized: %
\begin{equation}
\begin{aligned}
    \mathcal{L}(f_{\langle \theta,\phi,\eta \rangle}) =  - &\alpha \cdot \mathbb{E}_{Z \sim Q_\theta} \left[ \log{P_\phi}(X|Z,Y,\bar{Y})\right] \\  
    + &\beta \cdot D_{KL} (Q_\theta(Z|X) || P_\phi(Z)) \\ 
    - &\gamma \cdot \mathcal{L}_{Disc}(f_\eta)
    .
   \end{aligned}
\end{equation}
The above loss function takes into consideration both the quality of the reconstructed data 
and the ability to conceal private attributes. 
Thus, optimizing this loss function minimizes the weighted combination of 
the reconstruction loss and KL divergence term to 
encourage the reconstructed data segment to maximally retain non-private information. %
Meanwhile, the discriminator loss is maximized to ensure an adversary cannot 
infer private information from the latent representations, 
boosting Blinder's anonymization capability.
Note $\alpha$, $\beta$, and $\gamma$ are three hyper-parameters 
that are tuned empirically using grid search to balance privacy and utility. 
We elaborate the process of optimizing Blinder's loss function 
using a personalized federated learning algorithm in Section~\ref{subsec:meta_learning}.

When extending Blinder to protect multiple private attributes $Y_1, ..., Y_k$, 
additional discriminators can be introduced to 
ensure that $Z$ does not contain information about any of these attributes.
Specifically, the loss of these discriminators, i.e. $L_{Disc_1}, ..., L_{Disc_k}$,
are appended to Blinder's loss and optimized simultaneously.
Moreover, all these private attribute labels are sent to the decoder
along with the public attribute and $Z$.

\begin{algorithm}[t]
\small
\SetKwInOut{Execute}{Server Executes}
\KwData{Client set $\mathcal{S}$, no. clients selected for training $m$, meta-learning rate $\lambda$, local training dataset $\mathcal{D}_i$, local learning rate $\lambda_i$, support set size $s$, query set size $q$, communication rounds per epoch $t$} 
\KwResult{Blinder's parameters: $\{\theta, \phi, \eta\}$}
\Execute{}
\Indp
    Initialize $\{ \theta^1, \phi^1, \eta^1 \}$ \\
    \For{each epoch $c=1,2,…$} {
        $\mathcal{S}_c \leftarrow$ randomly select $m$ clients from $\mathcal{S}$ \\
        \For {each communication round $1, 2,…,t$} {
            \For{each client $i \in \mathcal{S}_c$ (in parallel)} {
                Send model $\{ \theta^t, \phi^t, \eta^t \}$ to client $i$ \\
                Receive updated gradients: $\{
                {\nabla_{\theta^t}\mathcal{L}(f_{ \langle {\theta'}_i^t, {\phi'}_i^t \rangle |{\eta'}_i^t})}$, 
                ${\nabla_{\phi^t}\mathcal{L}(f_{ \langle {\theta'}_i^t, {\phi'}_i^t \rangle |{\eta'}_i^t})}$,
                $\nabla_{\eta^t}\mathcal{L}_{Disc}(f_{{\eta'}_i^t|{\theta'}_i^t})\}$
            }
                \tcp{Update global model parameters}
                $\theta^{t+1} \leftarrow \theta^t - \frac{\lambda}{m} \sum^m_{i=1} \nabla_{\theta^t} \mathcal{L}(f_{ \langle {\theta'}^t_i, {\phi'}^t_i \rangle | {\eta'}^t_i})$\\
                $\phi^{t+1} \leftarrow \phi^t - \frac{\lambda}{m} \sum^m_{i=1} \nabla_{\phi^t} \mathcal{L}(f_{ \langle {\theta'}^t_i, {\phi'}^t_i \rangle | {\eta'}^t_i})$\\
                $  \eta^{t+1} \leftarrow \eta^t - \frac{\lambda}{m} \sum^m_{i=1} \nabla_{\eta^t} \mathcal{L}_{Disc}(f_{{\eta'}^t_i | {\theta'}^t_i})$ \\
        }

        }
\Indm
\SetKwInOut{Execute}{Client Executes}
\Execute{}
\Indp
     Pulling $\theta, \phi, \eta$ from the server \\ 
        Sample support set $\mathcal{D}_{i_s}$ and query set ${\mathcal{D}_{i_q}}$ from $\mathcal{D}_i$ \\
        $\mu(X_{i_s};\theta), \sigma(X_{i_s};\theta) \leftarrow$ $f_\theta(X_{i_s})$ \\
        $Z_{i_s} \leftarrow \mu(X_{i_s};\theta) + \sigma(X_{i_s};\theta) \odot \epsilon, \epsilon \sim \mathcal{N}(0,I)$  \\
        $\widetilde{X}_{i_s} \leftarrow$ $f_\phi(Z_{i_s} \oplus Y_{i_s} \oplus \bar{Y}_{i_s})$ \\
        $P(Y_{i_s}|Z_{i_s}) \leftarrow$ $f_\eta(Z_{i_s})$ \\
        \tcp{Adapt models on support set $\mathcal{D}_{i_s}$}
        $\eta' \leftarrow \eta - \lambda_i \nabla_{\eta} \mathcal{L}_{Disc}(f_{\eta|\theta})$ , with $\theta$ fixed\\
        $\langle \theta', \phi' \rangle \leftarrow \langle \theta, \phi \rangle  - \lambda_i \nabla_{\langle \theta, \phi \rangle} \mathcal{L}(f_{\langle \theta, \phi \rangle |\eta})$
        , with $\eta$ fixed \\
        \tcp{Compute gradients on query set $\!\!\mathcal{D}_{i_q}$}
        Estimate meta loss $\!\mathcal{L}(f_{\langle \theta', \phi' \rangle |\eta'}),\!\mathcal{L}_{Disc}(f_{\eta'|\theta'}\!)$ on ${\mathcal{D}_{i_q}}$ \\
        Compute gradients on meta loss w.r.t. $\{ \theta, \phi, \eta \}$:
        $\{
            {\nabla_{\theta}\mathcal{L}(f_{\langle \theta', \phi' \rangle|\eta'})}, 
            {\nabla_{\phi}\mathcal{L}(f_{\langle \theta', \phi' \rangle|\eta'})},
            \nabla_{\eta}\mathcal{L}_{Disc}(f_{\eta'|\theta'})\}$ \\
        Send gradients to the server\\
\Indm
 \caption{Distributed training of Blinder} 
 \label{alg:central_meta_obscurenet}
\end{algorithm}

\section{Distributed Training of Blinder}
We first outline the personalized federated learning algorithm 
where a central server coordinates 
a number of devices so that they can jointly train Blinder on their local data.
We then explain how fully trained Blinder can be used to anonymize 
sensor data on each device.

\subsection{Meta-Learned Anonymization Model}
\label{subsec:meta_learning}
\begin{figure}[tb]
\centering
\includegraphics[width=0.9\linewidth]{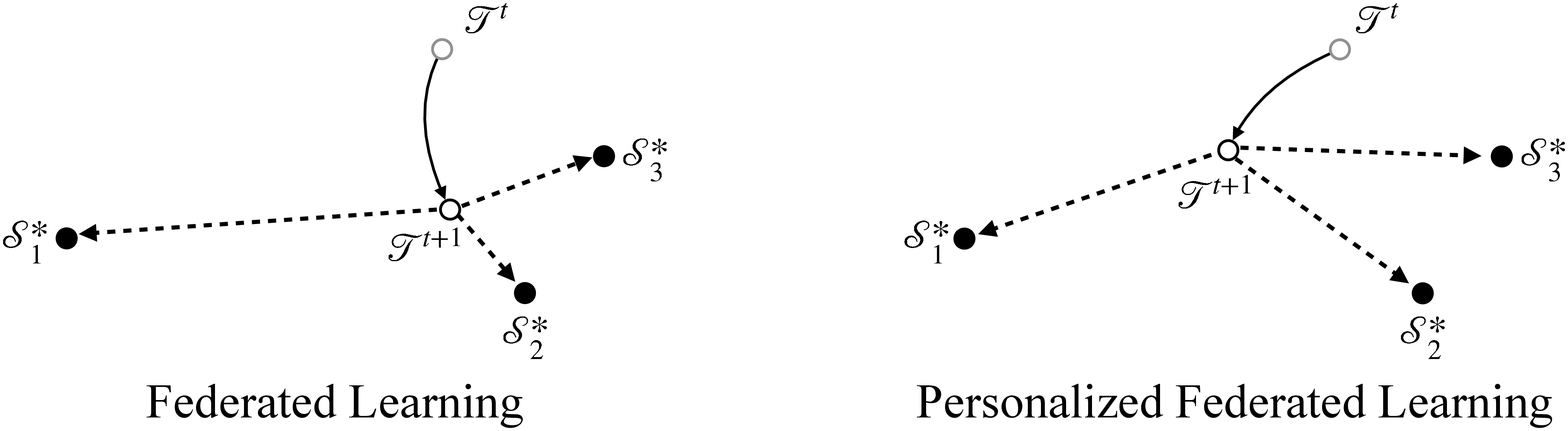}
\caption{Different aggregation methods used in FL.
The arrow between $\mathcal{T}^t$ and $\mathcal{T}^{t+1}$ indicates the update direction of the main task in iteration $t$. Arrows labeled $\mathcal{S}_1^*$, $\mathcal{S}_2^*$, and $\mathcal{S}_3^*$ show the optimal direction for sub-task $\mathcal{S}_1$, $\mathcal{S}_2$, and $\mathcal{S}_3$, respectively.
}
\vspace{-4mm}
\label{fig:agg_both}
\end{figure}

We consider learning the parameters of a global anonymization model $\mathcal{M}$ 
on training data residing on $N$ mobile/edge devices as our main learning task $\mathcal{T}$. 
This task is divided into $N$ sub-tasks: $\mathcal{S}_1, \mathcal{S}_2, ..., \mathcal{S}_N$,
where each sub-task entails learning a local anonymization model $\mathcal{M}_i$ on $\mathcal{U}_i$'s local data. 
This allows us to align the definition of tasks in our problem with meta-learning,
and consequently apply meta-learning to federated learning. 
In federated learning, the main task $\mathcal{T}$ is learned 
by taking the average of the parameters of local models $\mathcal{M}_i$ via FedAvg~\cite{mcmahan2017communication}.
In a classification task, FedAvg could cause the global model 
to be biased toward the majority classes that have more data samples, 
deviating from the main task of training a model that performs well for all classes.
To address this issue, we learn $\mathcal{M}$ through meta-learning. 
The meta-learned task differs from the task in FL in that 
our global model is optimized towards the direction that could quickly adapt to all sub-tasks. 
We illustrate this difference in Figure~\ref{fig:agg_both}, 
where the arrow between $\mathcal{T}^t$ and $\mathcal{T}^{t+1}$ is the update direction of the main task in iteration $t$,
and $\mathcal{S}_i^*$ is the optimal update direction %
for each sub-task $\mathcal{S}_i$.
Assuming that the local data in sub-tasks $\mathcal{S}_2$ and $\mathcal{S}_3$ correspond to a non-minority class, the update direction of the main task could be biased towards these classes, resulting in poor performance for the minority class, which exists in the local data of subtask $\mathcal{S}_1$. In personalized FL, the update direction of the main task is computed such that it enables better generalization for all sub-tasks.

Algorithm~\ref{alg:central_meta_obscurenet} is our personalized federated learning algorithm.
In each global epoch of this federated learning algorithm, the central server randomly 
selects a subset of $m\leq N$ clients from the $N$ available clients to participate in the model training.
We assume a selected client will remain available in this epoch and contribute 
to the model aggregation in multiple rounds of communication with the server.
At the beginning of each round, 
the server broadcasts a copy of the global anonymization model to each selected client (device).
Each selected client $\mathcal{U}_i$ then performs a two-step meta-learning optimization 
based on its local dataset $\mathcal{D}_i$. 
In the first step, $\mathcal{U}_i$ updates their local model on a small portion of data of size $s$, 
sampled from $\mathcal{D}_i$.
We call this small training batch, the \emph{support set} 
according to the conventions of few-shot learning, %
and denote the support set of $\mathcal{U}_i$ 
by $\mathcal{D}_{i_s}=\{ X_{i_s}, Y_{i_s}, \bar{Y}_{i_s} \}$.
The local encoder parameterized by $\theta_i$ takes as input the sensor data segment $X_{i_s}$ 
and maps it %
into a latent representation $Z_{i_s}$. 
The latent representation $Z_{i_s}$ serves two purposes.
First, $Z_{i_s}$, $Y_{i_s}$ and $\bar{Y}_{i_s}$ are concatenated and fed 
to the decoder to reconstruct the anonymized data $\widetilde{X}_{i_s}$.
Second, the discriminator, playing the role of the adversary, 
tries to predict the private attribute given $Z_{i_s}$.
Hence, the training process for the discriminator and VAE forms an adversarial minimax game, 
as discussed in Section~\ref{subsec:deep_generative_model_for_data_anonymization}. 
The parameters of the local discriminator $\eta_i$ are first optimized through backpropagation, 
while the parameters of the encoder are %
fixed (we write $\mathcal{L}_{Disc}(f_\eta)$ as $\mathcal{L}_{Disc}(f_{\eta|\theta})$ in Algorithm~\ref{alg:central_meta_obscurenet} to emphasize this). 
Then, we fix $\eta_i$ to optimize the parameters of the encoder and decoder, i.e., $\theta_i$ and $\phi_i$ 
(hence we write $\mathcal{L}(f_{\langle \theta,\phi,\eta \rangle})$ as $\mathcal{L}(f_{ \langle \theta,\phi \rangle |\eta})$).
The updated model parameters $\theta', \phi', \eta'$ are learned 
from the support set to optimize the client's local sub-task $\mathcal{S}_i$. 
The pseudocode for the first step optimization is shown in Algorithm~\ref{alg:central_meta_obscurenet} 
from Line~16 to~21. 
For better convergence, the first-step optimization can be repeated multiple times, 
each time is a local training step.

In the second step, we sample another batch of $\mathcal{U}_i$'s local data ${\mathcal{D}_{i_q}}$,
namely the \emph{query set}, 
where $\mathcal{D}_{i_s} \cap {\mathcal{D}_{i_q}} = \varnothing$,
to compute the meta loss $\mathcal{L}(f_{\langle \theta',\phi' \rangle})$ and $\mathcal{L}_{Disc}(f_{\eta'})$ 
of the model updated on the support set in the first step.
The meta loss is computed on a query set that is distinct from the support set 
for better generalization and enhanced performance on unseen data. %
By computing the gradients on the meta loss with respect to 
the original global anonymization model $\theta, \phi, \eta$, 
the global anonymization model can directly acquire more general knowledge of sub-task $\mathcal{S}_i$.
Hence, the meta-learned local losses yield an optimization direction that supports better adaptation for all sub-tasks.
The local meta losses can be passed to the central server to perform loss aggregation, 
then a backpropagation step is performed on the aggregated loss to update $\theta, \phi, \eta$. 
However, this approach requires sharing local computational graphs 
with the central server for backpropagation, introducing additional communication overhead.
For a more efficient implementation, 
we take an alternative approach based on Federated~SGD~\cite{mcmahan2017communication}, 
to first compute the gradients on the meta losses 
with respect to the global model parameters $\theta, \phi, \eta$ on the client side, 
then send the gradients (rather than the meta losses) back to the central server.
The second-step optimization is summarized in Algorithm~\ref{alg:central_meta_obscurenet} from Line~22 to~24.
    Note that we explore the feasibility of data anonymization under distributed settings following a standard FedSGD.
    Applying more advanced aggregation methods, 
    such as~\cite{li2020federated,wang2020tackling}, to Blinder 
    would be a straightforward extension, which is not discussed in this work.

After receiving the gradients from a sufficient number of clients, 
the server updates the parameters of the global anonymization model $\theta$, $\phi$, $\eta$ 
through a weighted aggregation.
The update rules are from Line~9 to~11, %
where $\lambda$ is the meta learning rate,
$\nabla_{\langle \theta^t, \phi^t \rangle}\mathcal{L}(f_{ \langle {\theta'}^t_i, {\phi'}^t_i \rangle | {\eta'}^t_i})$ and $\nabla_{\eta^t}\mathcal{L}_{Disc}(f_{{\eta'}^t_i | {\theta'}^t_i})$ are
the gradients computed by the clients. %

\subsection{Anonymization Process}\label{sec:anon-process}
Upon obtaining the trained Blinder, clients can retrain this model on their local data 
(as an optional adaptation step) and then use it to anonymize their sensor data 
by predicting the corresponding public attribute using a pre-trained machine learning model 
and manipulating the private attribute, both of which are appended to the latent representation $Z$ 
before feeding it to the decoder.
While it is possible to change the private attribute using a fixed bijective function, 
this anonymization process does not mitigate the \emph{re-identification attack}
as discussed in~\cite{hajihassnai2021obscurenet}.
To put it simply, if we always change private attribute class $i$ to private attribute class $j$,
the adversary can easily learn this mapping and foil the data anonymization.
To address this problem, we adopt a stochastic anonymization approach, 
which draws a private attribute in a uniformly random manner 
from the set of values the private attribute can take and pass it %
to the decoder to obscure the data samples.
This approach only requires the prior 
knowledge of the set of possible private attribute classes. %

\section{Addressing Heterogeneity}\label{sec:rebalance}
We present the key challenges in handling imbalanced class distributions in FL.
These challenges are addressed by utilizing re-sampling techniques and Blinder's decoder 
to balance the distribution of public and private attributes in each user's device, respectively.  

\subsection{Public Attribute Distribution}
\label{subsec:balance_public}

In supervised learning, the model trained on an imbalanced dataset, where classes are not represented equally,
would be biased towards the majority class (i.e., the class with most samples) 
and perform poorly for samples that belong to minority classes (i.e., classes with fewer samples).
This problem is exacerbated in federated learning because the majority class 
could be different across the clients that collaboratively train a model, 
creating a challenging non-i.i.d. data situation~\cite{wang2020tackling, li2020federated}.
Specifically, training Blinder on distributed sensor data via personalized federated learning  
forces the global model to be updated according to the aggregate of heavily-biased local gradients
when the distributions of public attributes vary significantly among the clients.
This hampers the performance of Blinder.
To address this issue, we balance the distribution of public attributes
via a combination of over-sampling and under-sampling techniques.

In this work, we say a dataset is not balanced when a majority class 
contains at least twice as many data samples as the minority class.
To increase the number of data samples from the minority classes, 
simply oversampling from the existing dataset could produce many duplicate samples and cause overfitting.
Instead of doing that, we leverage SMOTE~\cite{chawla2002smote} 
to generate unique, high-quality data samples to supplement minority public attributes. 
SMOTE randomly draws samples from the minority classes and finds their $K$ most similar neighbors 
among the real samples. 
A feature space is then generated from the selected sample and its $K$ nearest neighbors, 
from which the synthesized data will be sampled.
We evaluate the quality of the synthesized data using
public and private attribute classifiers, 
both of which are pre-trained on the original data. 
We find that these classifiers identify the public and private attributes of
the synthesized data with $96\%$ to $98\%$ accuracy. %
Moreover, we resort to random under-sampling of the majority class
to better balance the distribution when the number of data samples in the majority class 
is several orders of magnitude higher than that of the minority classes.

\subsection{Private Attribute Distribution}
\label{subsec:balance_private}
\begin{figure}[t]
    \centering
    \begin{minipage}{0.48\linewidth}
        \centering
        \includegraphics[width=0.98\linewidth]{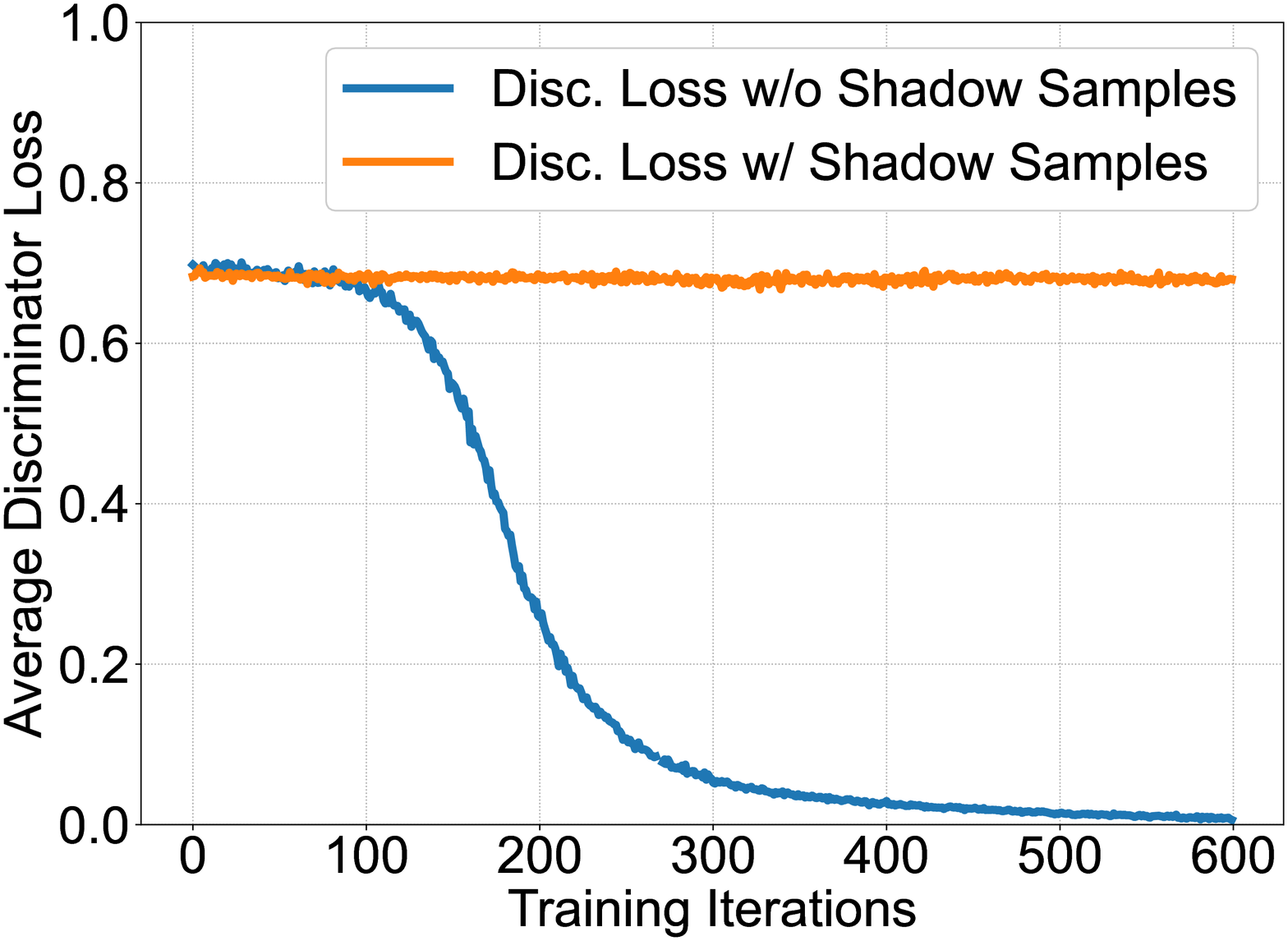}
        \vspace{-1mm}
        \caption{Average discriminator loss %
        computed on the query set w/ and w/o shadow samples.}
        \label{fig:disc_loss}
    \end{minipage}%
    \hfill
    \begin{minipage}{.48\linewidth}
        \centering
        \vspace{3mm}
        \includegraphics[width=\linewidth]{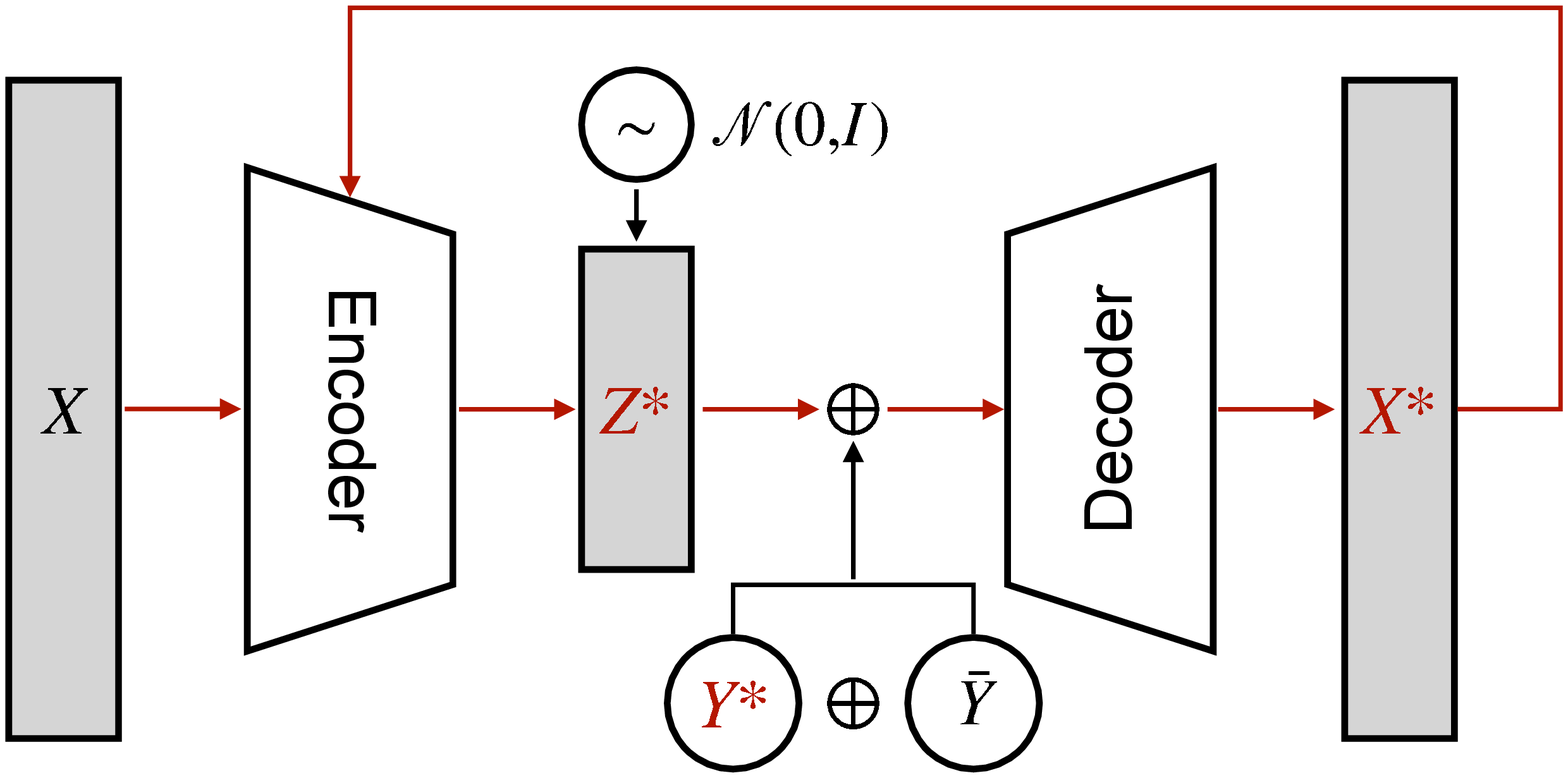}
        \vspace{2mm}
        \caption{Generation of shadow samples using Blinder's VAE architecture. %
        Data flow is shown in red.}
        \label{fig:synthetic_sample}
    \end{minipage}
    \vspace{-1mm}
\end{figure}

Consider a practical, yet challenging case in which the user's private attribute does not vary with time 
(the assumption made in Section~\ref{sec:problem_def}). %
This creates a non-trivial overfitting issue in~FL
and significantly degrades Blinder's privacy-preserving capability. %
More precisely, when a user's local dataset contains only one private attribute class,
the local discriminator becomes unusable as it will quickly overfit and 
predict the correct private attribute regardless of how the latent representation is extracted. %
Figure~\ref{fig:disc_loss} depicts a drastic decline 
in the local discriminator loss (the blue curve) as a result of overfitting.
Recall that in the minimax game between the discriminator and encoder,
the discriminator must provide useful feedback to the encoder 
to facilitate learning a latent representation that contains no information about the private attribute. 
An overfitted discriminator cannot provide this feedback, 
preventing the encoder from learning how to conceal the private attribute in the latent representation.
With multiple local anonymization models failing to perform well in their respective tasks, 
the anonymization capability of the aggregated global anonymization model would degrade too.

The personalized federated learning algorithm can partially 
mitigate this issue compared to the standard federated learning algorithm 
thanks to the application of meta-learning.
To fully address this issue, we augment each user's local data
by synthesizing samples that contain information 
about other possible private attribute classes. 
This can diversify the private attribute in each user's dataset, 
thereby avoiding overfitting of the local discriminator.

We denote the synthesized data samples containing new private attributes by $X^*$, 
and call them \emph{shadow samples} because they have the same feature distribution as the original data samples,
yet a different private attribute than this user's actual private attribute.
We utilize the feed-forward propagation of our VAE architecture to generate the shadow samples 
as depicted in Figure~\ref{fig:synthetic_sample}.
The first step is to feed original sensor data $X$ to the encoder 
to obtain the latent representation %
denoted $Z^*$. 
The reparameterization in~(\ref{eq:reparam}) ensures that $Z^*$ is sampled randomly 
from the latent space distribution.
Next, we pick a private attribute $Y^*$, 
which can be any private attribute class except the user's real private attribute $Y$. %
Having been learned collaboratively by a diverse set of users, 
Blinder's decoder can reconstruct the data segment given any condition, i.e., private attribute class. 
Hence, we generate the shadow samples by feeding $Z^* \oplus Y^* \oplus \bar{Y}$ to the decoder.
The shadow samples are then encoded and their representations are mixed with other representations 
in the user's original support set to train the local discriminator.
The orange curve in Figure~\ref{fig:disc_loss} shows the discriminator loss 
on the query set averaged over all users after incorporating the shadow samples. 
It can be seen that the discriminator loss fluctuates near the random guess level 
($0.69$ for the binary cross-entropy loss), 
suggesting that adversarial training works properly.
The main advantage of our approach is that it simply uses Blinder's VAE architecture 
for augmenting the user's dataset without relying on an additional generative model.
Besides, it only requires prior knowledge of the set of possible private attribute labels, 
which is %
consistent with the assumption made for stochastic anonymization in Section~\ref{sec:anon-process}.
By incorporating the shadow samples, 
Blinder further reduces the overall accuracy of intrusive inferences by around ${5\%}$. 

\begin{figure}[t]
    \centering
    \begin{subfigure}[t]{0.49\linewidth}
         \centering
         \includegraphics[width=\linewidth]{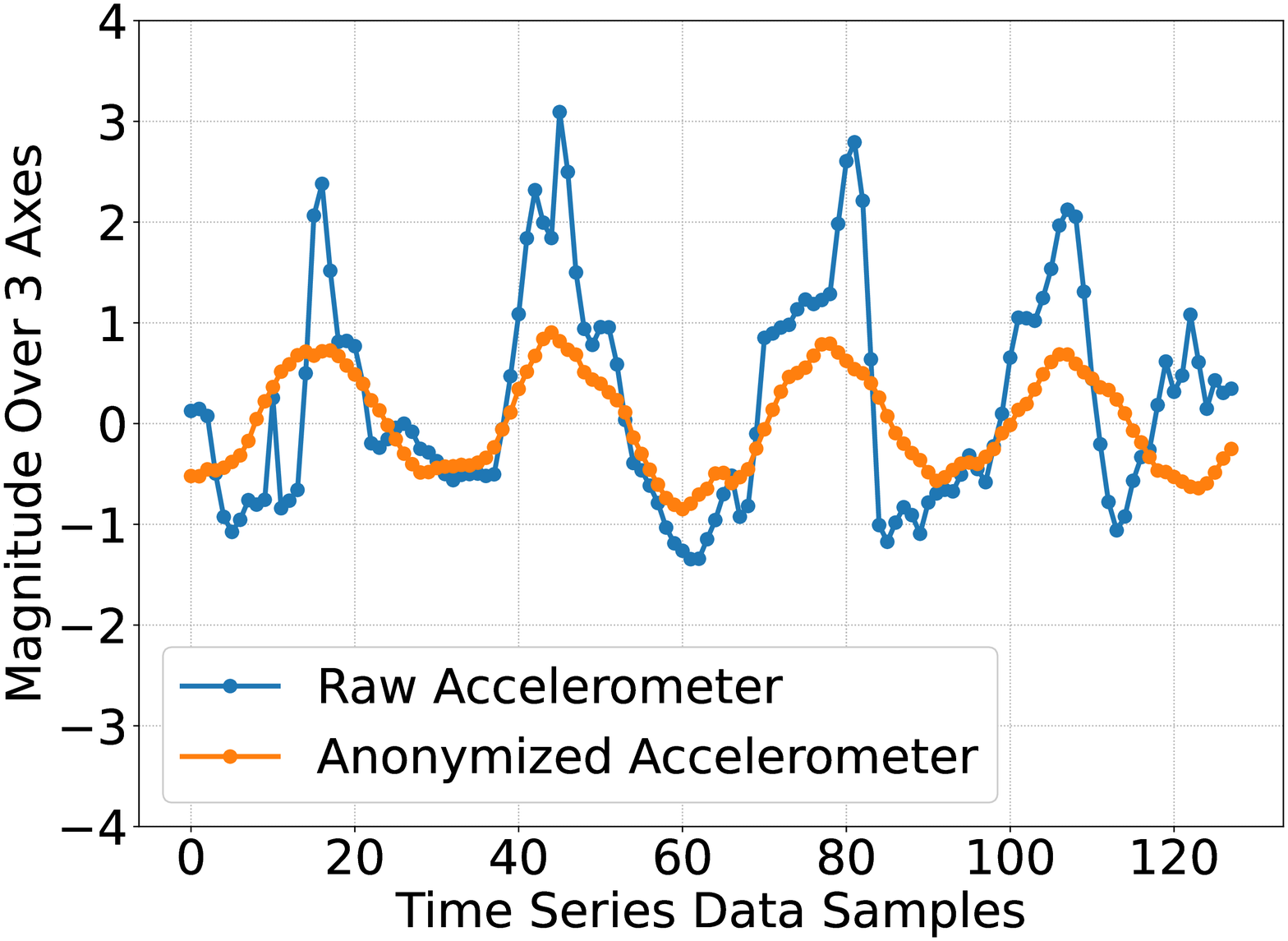}
         \caption{Anonymizing accelerometer data}
         \label{fig:signal_acc}
     \end{subfigure}
     \hfill
     \begin{subfigure}[t]{0.49\linewidth}
         \centering
         \includegraphics[width=\linewidth]{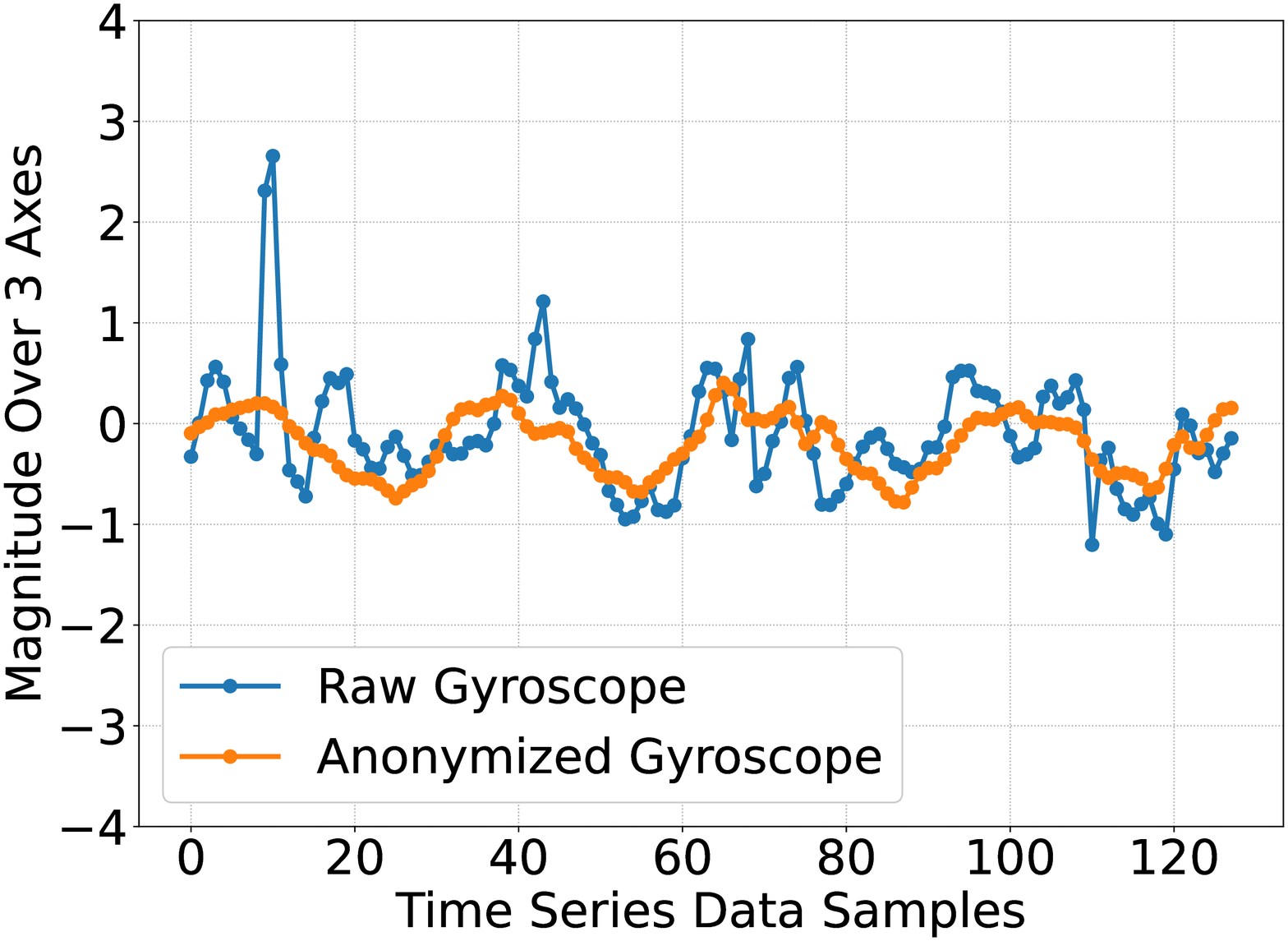}
         \caption{Anonymizing gyroscope data}
         \label{fig:signal_gyro}
     \end{subfigure}
     \vspace{-2mm}
    \caption{Comparison between the original sensor readings and the anonymized sensor readings on the MotionSense dataset.}
    \vspace{-4mm}
    \label{fig:signals}
\end{figure}

We use the motion sensor data in MotionSense as an example (details provided in Section~\ref{subsec:datasets}) to compare the original sensor readings with the reconstructed readings by Blinder. As illustrated in Figure~\ref{fig:signals}, for both accelerometer and gyroscope readings, Blinder can effectively learn the salient motion patterns from the magnitude of IMU signals to maintain the activity recognition accuracy. On the other hand, the high-frequency components of the motion data are smoothed out as they can reveal the identity or other characteristics (like body size) of the individual. Notice that the magnitude of original readings fluctuates in a wider range than anonymized readings. We argue that it is especially helpful to protect certain private attributes. For instance, the magnitude of the signal has a broader range for taller/male individuals due to the larger movements they might make.

\section{Evaluation}\label{sec:eval}

We compare Blinder with four %
baselines described below:
\subsubsection*{Baseline 1 - Anonymization Autoencoder (AAE)~\cite{malekzadeh2019mobile}:}
    It is an autoencoder-based data anonymization mode that can serve as an interface between raw sensor data and untrusted data consumers.
    Unlike Blinder which conditions the latent information and leverages generative models to conceal private attributes, 
    AAE employs multiple regularizer modules based on neural networks to control the information learned by the autoencoder. 
    The encoder module of AAE comprises four stacked convolutional layers, each followed by a batch normalization layer. 
    The decoder module of AAE reverses the encoding process using transposed convolutional layers and is also batch normalized.
    AAE requires to be trained on a centralized server and hence can serve as the centralized baseline to compare with distributed anonymization models.
    We implement AAE based on the source code released by the authors on GitHub~\cite{aae-git}. 
    Note that AAE uses a deterministic anonymization strategy which makes it 
    vulnerable to the re-identification attack discussed in~\cite{hajihassnai2021obscurenet}.

\subsubsection*{Baseline 2 - ObscureNet~\cite{hajihassnai2021obscurenet}:}
    It is a data anonymization model based on the VAE architecture.
    It is shown in~\cite{hajihassnai2021obscurenet} that it has better privacy-preserving
    performance than several other data anonymization models that utilize autoencoders. %
    This model is trained on a server that hosts centralized training data; 
    thus, it requires users to send their raw sensor data 
    along with the corresponding public and private attributes to this server. 
    Unlike Blinder, which conditions the latent feature on both private attribute $Y$ and public attribute $\bar{Y}$, ObscureNet only sends the private attribute to the decoder. 
    Hence, it cannot generalize to all public attributes and 
    requires a dedicated anonymization model per public attribute.
    We used the source code published by the authors on GitHub~\cite{obscurenet-git}, 
    and considered the version of ObscureNet that performs stochastic anonymization as our baseline.
    Since ObscureNet is trained in a centralized fashion, 
    comparing it with Blinder will help to understand  
    how %
    training the anonymization model on decentralized data affects utility and privacy loss.

    \subsubsection*{Baseline 3 - FedBlinder:}
    We implement the third baseline, named FedBlinder,
    by training the three neural networks in Blinder via federated learning without applying meta-learning. 
    In this case, a central server randomly selects $m$ users from the set of available users in each epoch 
    to participate in model training and sends them a copy of the global anonymization model.
    Each selected user retrains the received copy of the global model on their 
    device for $5$ rounds using their local data. 
    When the local training process completes, the updated model parameters (rather than the gradients) 
    will be transmitted to the central server for FedAvg~\cite{mcmahan2017communication}-based aggregation.
    In FedBlinder, we do not apply the rebalancing techniques proposed in Section~\ref{sec:rebalance}, 
    nor do we take advantage of meta-learning for fast adaptation.
    Therefore, the comparison between Blinder and FedBlinder 
    will highlight the significance of the proposed personalized federated learning and rebalancing techniques.

    \subsubsection*{Baseline 4 - KD-Blinder:}
        This baseline extends the FedBlinder baseline
        by taking advantage of the knowledge distillation (KD) techniques 
        proposed in BalanceFL~\cite{shuai2022balancefl} 
        to tackle missing class and class imbalance issues in federated learning.
        Specifically,
        BalanceFL utilizes a knowledge inheritance technique and incorporates a smooth regularization term to tackle the missing class issue. 
        Moreover, it uses inter-class balancing techniques, including class-balanced sampling and feature-level data augmentation, to balance the training data.
        Since in our data anonymization task, the private attribute suffers from the missing class issue and the public attribute suffers from the class imbalance issue, 
        we apply the knowledge inheritance and smooth regularization techniques to address the missing class issue during the auxiliary model training, and use the class-balanced sampling technique and feature-level data augmentation to address the class imbalance issue when training the CVAE model. 
        We implement these techniques following the source code released by the authors on GitHub~\cite{balancefl-git}.
        Note that in BalanceFL~\cite{shuai2022balancefl}, the feature-level data augmentation is performed on the output of the second last layer in a classifier, 
        but our CVAE model is not a classifier. 
        Therefore, we perform it on the latent representation generated by Blinder's encoder.
        The comparison between KD-Blinder and Blinder will show 
        the efficacy of the public attribute rebalancing technique and 
        shadow sample synthesis that we use in Blinder to address the missing class issue 
        for the private attribute.

\subsubsection*{Implementation Details:}
We implement Blinder,
FedBlinder, KD-Blinder, and ObscureNet using PyTorch~\cite{NEURIPS2019_9015}. 
The AAE baseline is implemented using Keras~\cite{chollet2015keras} 
based on the code~\cite{aae-git} released by the authors.
The personalized federated learning algorithm is implemented on top of the learn2learn library~\cite{Arnold_learn2learn_A_Library_2020}. 
In each round of training, we randomly select $40\%$ of all users to participate in that round.
We assume the selected users will stay connected to the central server (i.e., no stragglers\footnote{Dealing with stragglers and connectivity issues is important for the real-world deployment of FL algorithms.
However, we ignore these factors here due to the page limit and 
the fact that they have been addressed in the past~\cite{li2020federatedsurvey, chen2020joint}.})
until the last round of communication finishes. 
The local models are trained and averaged on a single server 
by different processes to simulate a federated learning setting, 
where the server will wait until all participants send the locally updated models before it aggregates them. 
The training batch size is $b=s+q=16$ as we do few-shot learning, 
with $s=1$ being the size of the support set and $q=15$ being the size of the query set. 
The dimension of the latent space is set to $25$.
We use grid search to tune the hyper-parameters: $\alpha=0.9$, $\beta=2$, $\gamma=0.2$.
Note that for the experiments conducted on MotionSense (described below), 
we change the first (last) two layers of the encoder (decoder) in Blinder and FedBlinder 
from fully connected layers to convolutional (transposed convolutional) layers for better convergence.
For a fair comparison with the baselines,
we do not perform local adaptation on the fully trained Blinder, 
except in Section~\ref{sec:adaptation} where it is discussed.

\subsection{Datasets}
\label{subsec:datasets}

We evaluate Blinder on three publicly available human activity recognition (HAR) datasets 
that include different sensing modalities. 
The first two datasets, namely MobiAct~\cite{chatzaki2016human} and MotionSense~\cite{malekzadeh2019mobile}, are collected by IMUs embedded in mobile devices. 
The third dataset~\cite{baha2020dataset} contains radio frequency signals. 
This dataset, which we call the Wi-Fi HAR dataset, 
utilizes the Wi-Fi channel state information (CSI).
\subsubsection*{MobiAct:}
This dataset contains sensor readings pertaining to 66 participants that perform 12 different daily activities,
captured by a 3-axis accelerometer, a 3-axis gyroscope, and an orientation sensor 
embedded in a Samsung Galaxy S3 smartphone~\cite{chatzaki2016human}. 
We select the same users and activities that were used
in the ObscureNet baseline~\cite{hajihassnai2021obscurenet} for a fair comparison.
We use the sensor readings of the accelerometer and gyroscope for 4 daily activities: 
walking, standing, jogging, and walking up the stairs, 
performed by 36 individuals, 20 of them are male and 16 are female.
We standardize and segment timeseries generated by these sensors 
using a sliding window of 128 samples and a stride length of 10 samples. 
The data segments of 6 sensor channels are then concatenated 
to create a one-dimensional tensor of size 768, which is fed to the anonymization models. 
Unless otherwise mentioned, the train-test split ratio is 8:2.
The public attribute classes are the 4 activities described above. 
We use Blinder to anonymize two distinct private attributes, 
namely the user's (binary) gender and (ternary) weight group attributes. 
We categorize a user's weight into one of the three weight groups, 
where group 0 is for weight below 70 kg, group 1 is for weight within 70 kg to 90 kg, 
and group 2 is for weight above 90 kg. 
This binning strategy divides users into three groups that have roughly the same size, 
allowing us to model the weight as a categorical variable %
that can be anonymized by Blinder.

\subsubsection*{MotionSense:}
This dataset contains readings of accelerometer and gyroscope sensors in an iPhone~6s
which is kept in the front pocket of 24 individuals (14 male and 10 female) who performed 6 different activities: 
walking up the stairs, walking down the stairs, walking, standing, jogging, and sitting. 
We only consider 4 of the 6 daily activities, namely walking up and down the stairs, walking, and jogging.
The sitting and standing activities are excluded because there are only a few samples for these activities,
and IMU sensor readings contain little information to distinguish between them when the individual is not moving.
Every individual performs 15 trials for each activity. 
For the sake of comparison with the baseline~\cite{hajihassnai2021obscurenet},
we focus on gender anonymization only and use the first 9 trials to create the training set. 
The remaining 6 trials are used in the test set.
We compute the magnitude over the 3 axes of each sensor and treat it as the sensor reading. 
We adopt the same segmentation technique that was described for MobiAct.

\subsubsection*{Wi-Fi HAR:}
    This dataset utilizes the Wi-Fi channel state information (CSI) 
    to perform human activity recognition~\cite{baha2020dataset}.
    The CSI data is collected using $1$ transmitter and $3$ receivers in $3$ environments. 
    Since each transceiver pair generates $30$ CSI channels, $90$ CSI channels are available in total.
    Two environments use the line-of-sight (LOS) configuration and 
    the third one has a non-LOS configuration. 
    A total of $30$ subjects participated in the experiments, 
    with each environment involving $10$ subjects. 
    Each participant performs $5$ types of experiments that include $12$ different activities; 
    each experiment is repeated $20$ times ($20$ trials).
    From this dataset, we select $4$ activities, namely standing, sitting, lying down, and turning around. We consider the weight and height of the participants as private attributes. 
    We use data from Environment~$2$ with the LOS configuration 
    because the participants' private attributes are more balanced. 
    By dividing the participants' weights and heights into $2$ groups 
    using a boundary of 80 kg and 175 cm, respectively,
    each weight and height group contains exactly 5 participants.
    To preprocess the data, we compute the signal magnitude of all available channels and perform data standardization, then we segment the CSI magnitude using a sliding window with a length of $80$ samples and a stride length of $40$ samples. 
    We randomly partition the dataset into the training set and test set with a ratio of $8{:}2$. 
    We use the same set of random seeds for data partitioning when repeating the experiments on Blinder 
    and all baselines to ensure fairness and reproducibility.

\subsection{Evaluation Metrics}
\label{subsec:eval_metrics}
\subsubsection{Impact on Privacy}
We evaluate the privacy-preserving capability of an anonymization model
by measuring the accuracy of intrusive inferences about the user-defined private attribute(s)
on the sensor data anonymized by this model.
In particular, we adopt a convolutional neural network (CNN) %
that consists of $4$ convolutional layers followed by $3$ fully connected layers
as our \emph{intrusive inference model}.
This intrusive inference model is pre-trained on raw sensor data as discussed in Section~\ref{sec:problem_def},
and its accuracy is a measure of privacy loss. 
This is because an HBC adversary can use the intrusive inference model
to de-anonymize the previously anonymized sensor data.
Thus, the minimum privacy loss is attained 
when the intrusive inference model's accuracy equals the level of random guessing.

\subsubsection{Impact on Data Utility}
\label{subsubsec:utility_evaluation_model}
We use the accuracy of desired inferences about the user-defined public attribute 
on the anonymized sensor data to quantify the utility of the anonymized data.
For example, in the HAR task, we use the accuracy of a powerful CNN-based activity recognition model,
which we call the \emph{desired inference model}, as a measure of data utility. %
The desired inference model has the same architecture as the intrusive inference model, 
and is pre-trained on raw sensor data as discussed in Section~\ref{sec:problem_def}.
We say that the maximum data utility is attained when the accuracy of the desired inference model 
remains the same after sensor data is anonymized.

\subsubsection{Generalizability}
\label{subsubsec:metric_hetero}
We assess how well an anonymization model generalizes 
to a heterogeneous population, i.e., non-i.i.d. data, and to new users.
First, we study the model's ability to generalize to users with different public attribute distributions.
To capture the skewness of the public attribute distribution in a user's local dataset,
we use an imbalance ratio $R_{D}$, which is defined as the number of samples 
in the class with most samples over the number of samples in the class with fewest samples. %
We assume all users share the same imbalance ratio, but may have different public attribute class distributions.
To this end, we randomly pick a subset of the public attribute classes from each user's local training set 
as the majority class(es), leaving the rest as minority classes. 
We then reuse the data synthesis and down-sampling techniques described in Section~\ref{subsec:balance_public},
this time to make the class imbalance issue more pronounced, reaching the target ratio $R_{D}$.
By varying $R_D$, we can evaluate Blinder and the baselines under settings with non-i.i.d. public attributes. 
Note that, for a fair comparison, we do not rebalance the public attributes as we normally do in Blinder.
Finally, we study the model's ability to generalize to users who never participated in model training. 
Let $R_U$ be the ratio of the users that participated in model training to the total number of users.
A model that generalizes well should perform well when trained on a small number of users (low $R_U$ ratio) 
and used to anonymize data of new users.

\begin{figure}[t]
    \centering
    \begin{subfigure}[b]{0.49\linewidth}
         \centering
         \includegraphics[width=\linewidth]{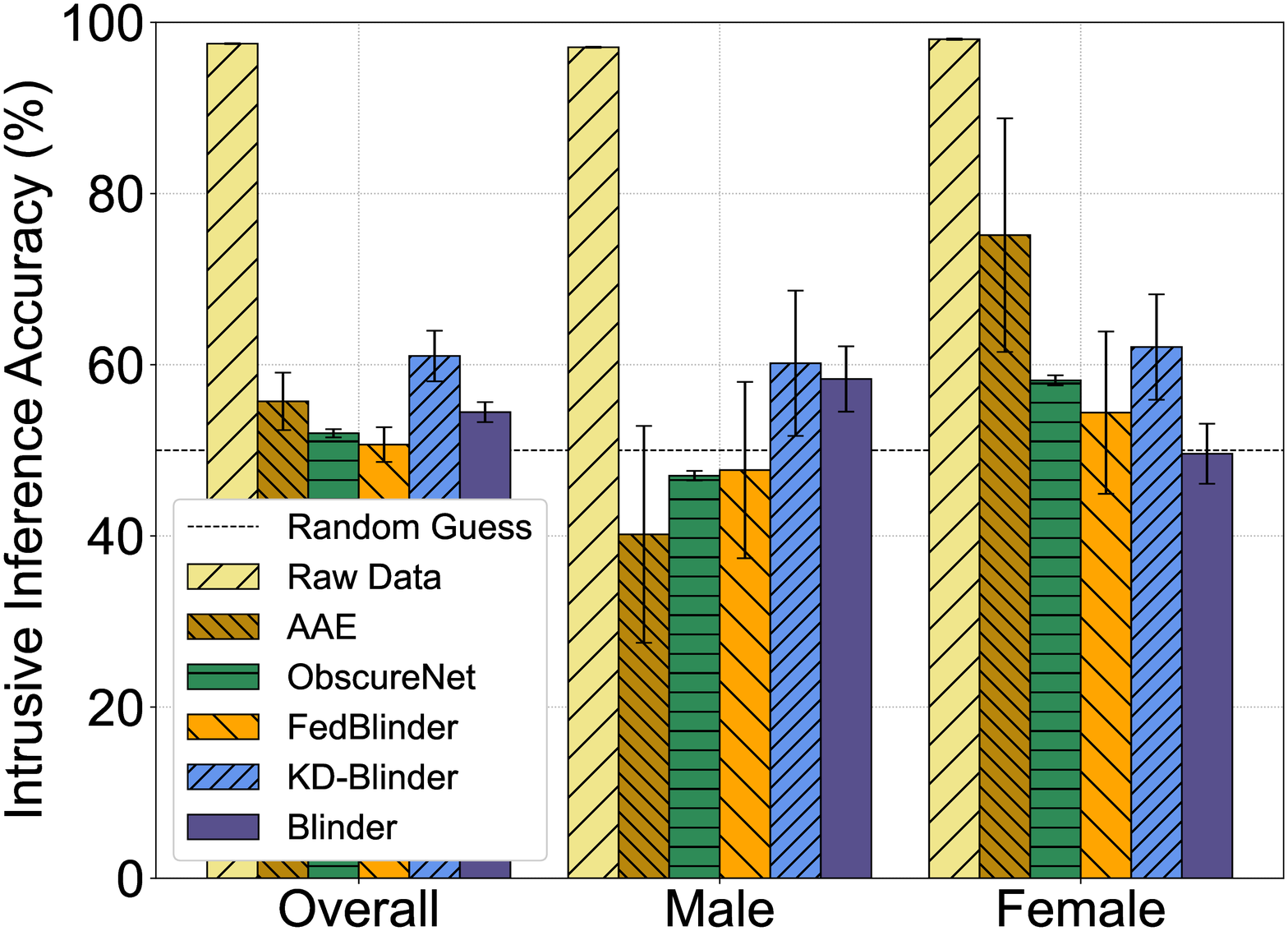}
         \caption{Gender anonymization}
         \label{fig:privacy_random_mobi_gen}
     \end{subfigure}
     \hfill
     \begin{subfigure}[b]{0.49\linewidth}
         \centering
         \includegraphics[width=\linewidth]{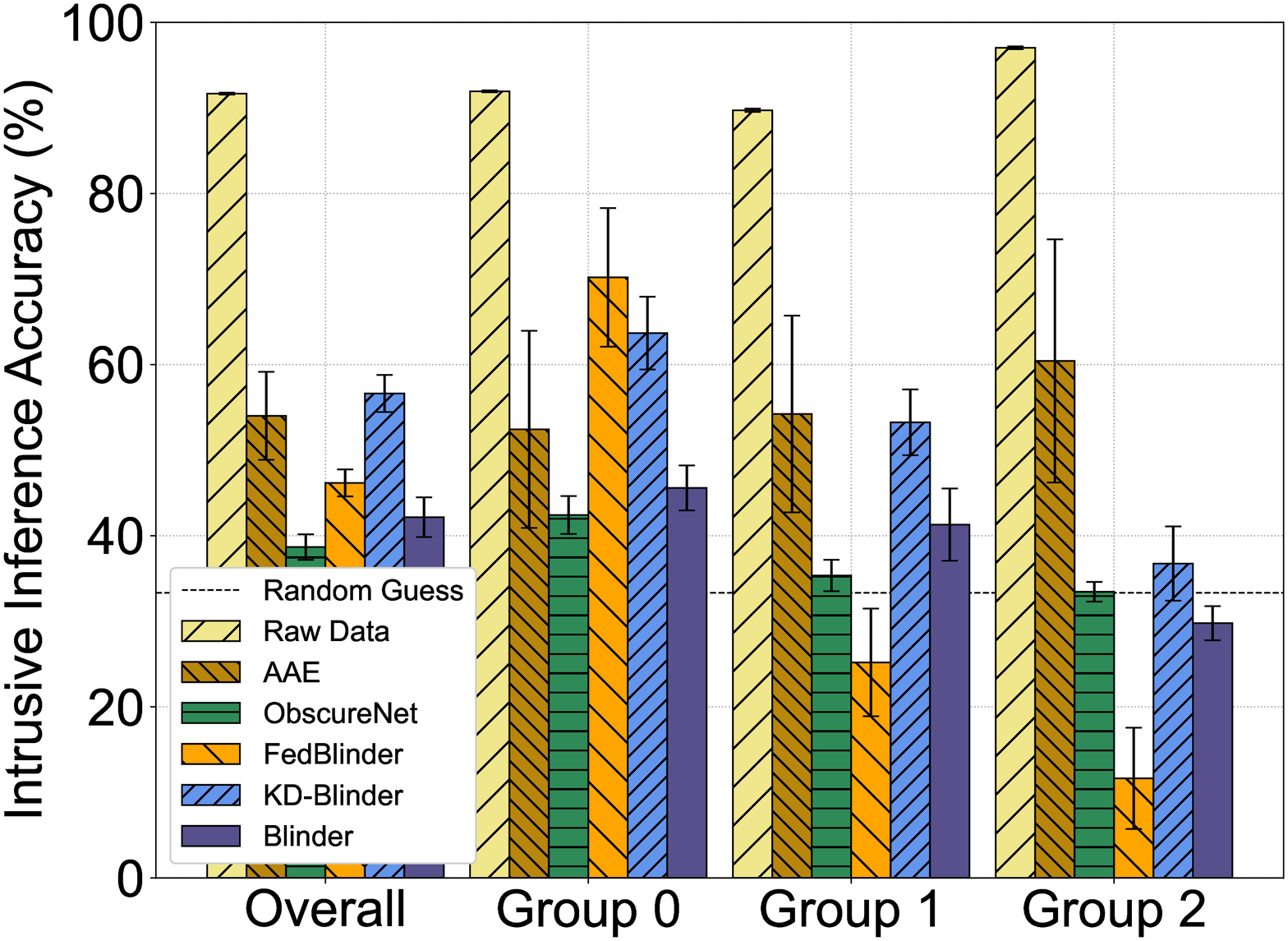}
         \caption{Weight anonymization}
         \label{fig:privacy_random_mobi_weight}
     \end{subfigure}
    \caption{Intrusive inference accuracy on MobiAct for gender and weight anonymization.}
    \label{fig:privacy_random_mobi}
\end{figure}

\subsection{Privacy-Preserving Performance}
\label{subsec:eval_privacy}
We first compare Blinder and the baselines in terms of their privacy-preserving capability on MobiAct.
Suppose the public attribute is the user's activity and the private attribute, which we aim to obscure, 
is their gender or weight group. 
We anonymize the sensor data and report the average accuracy of the intrusive inference model 
over 10 trials with different random seeds in Figure~\ref{fig:privacy_random_mobi}. 
The error bar shows the standard deviation across the 10 runs.
ObscureNet exhibits the best privacy-preserving performance among the five models, 
resulting in average intrusive inference accuracy of $51.99\%$ ($38.68\%$)
and F1 score of $51.99\%$ ($34.97\%$) in the gender (weight) anonymization task, 
reducing the intrusive inference accuracy by $43\%$ ($53\%$) compared to the case 
that the sensor data is not anonymized.
Despite its superb performance (being close to random guessing), 
ObscureNet must be trained in a server that stores the centralized training data.
AAE achieves average intrusive inference accuracy of $55.72\%$ ($54.01\%$)
and F1 score of $54.43\%$ ($50.88\%$) in the gender (weight) anonymization task.
Similar to ObscureNet, AAE is trained in a centralized fashion.
Hence, both ObscureNet and AAE do not
protect the privacy of users in the model training phase.
FedBlinder results in average intrusive inference accuracy of $50.67\%$ ($46.17\%$) and 
F1 score of $50.2\%$ ($34.4\%$) for gender (weight) anonymization task.
Although the average gender inference accuracy is slightly lower %
under FedBlinder than ObscureNet,
we see that the standard deviation of both gender and weight inference accuracy is noticeably higher (larger error bar)
under FedBlinder than both ObscureNet and Blinder.
We attribute this unstable performance to the inherent weakness of the vanilla FedAvg algorithm
when applied to datasets with highly imbalanced private attribute distributions, 
in this case the presence of just one gender/weight class in each local dataset.
This becomes worse when %
there are more private attribute classes, as is the case in the ternary weight anonymization task, where the average weight inference accuracy for users in weight group~0 is about $70.2\%$, which is much higher than its average accuracy among all three classes.
In particular, FedBlinder performs worst in the weight anonymization task 
and exhibits variable performance across different weight groups.
The average weight inference accuracy for users in weight group~0 is about $70.2\%$, 
which is much higher than its average accuracy across the three classes.
With the help of knowledge distillation and inter-class balancing techniques, KD-Blinder achieves a more balanced performance across the binary gender classes and the ternary weight classes than FedBlinder, yet it exhibits the worst privacy protection with an average accuracy (F1 score) of $61.02\%$ ($60.17\%$) and $56.62\%$ ($50.61\%$) for gender and weight anonymization, respectively.

Our proposed personalized federated learning framework allows Blinder 
to protect users' private attributes during model training, similar to FedBlinder and KD-Blinder.
That aside, we observe that Blinder shows a more stable privacy-preserving capability than both FL-based baselines, 
resulting in average intrusive inference accuracy of $54.46\%$ ($42.16\%$) 
and F1 score of $53.91\%$ ($37.47\%$) for gender (weight) anonymization. 
The gender (weight) inference accuracy shows a modest increase of around $2.47\%$ ($3.48\%$) compared to ObscureNet. 
Blinder outperforms FedBlinder in weight anonymization not only because of
the slightly better overall performance ($\sim{4\%}$), 
but also more consistent %
performance for each individual class.
The highest weight inference accuracy among the three weight groups is $45.58\%$, 
only $3.16\%$ higher than the highest weight inference accuracy under ObscureNet.
Although the knowledge inheritance technique used in KD-Blinder 
seems somewhat effective in dealing with missing classes, 
KD-Blinder has a lower anonymization capability than FedBlinder and significantly underperforms Blinder trained with the shadow samples. 
We believe this might be because the knowledge distillation technique 
prevents missing class labels from being updated during backpropagation.
This could restrict the learning capability of the auxiliary model 
and provide less useful feedback for the encoder 
to obscure information about the private attribute.

\begin{figure}[t]
    \centering
    \begin{subfigure}[b]{0.49\linewidth}
         \centering
         \includegraphics[width=\linewidth]{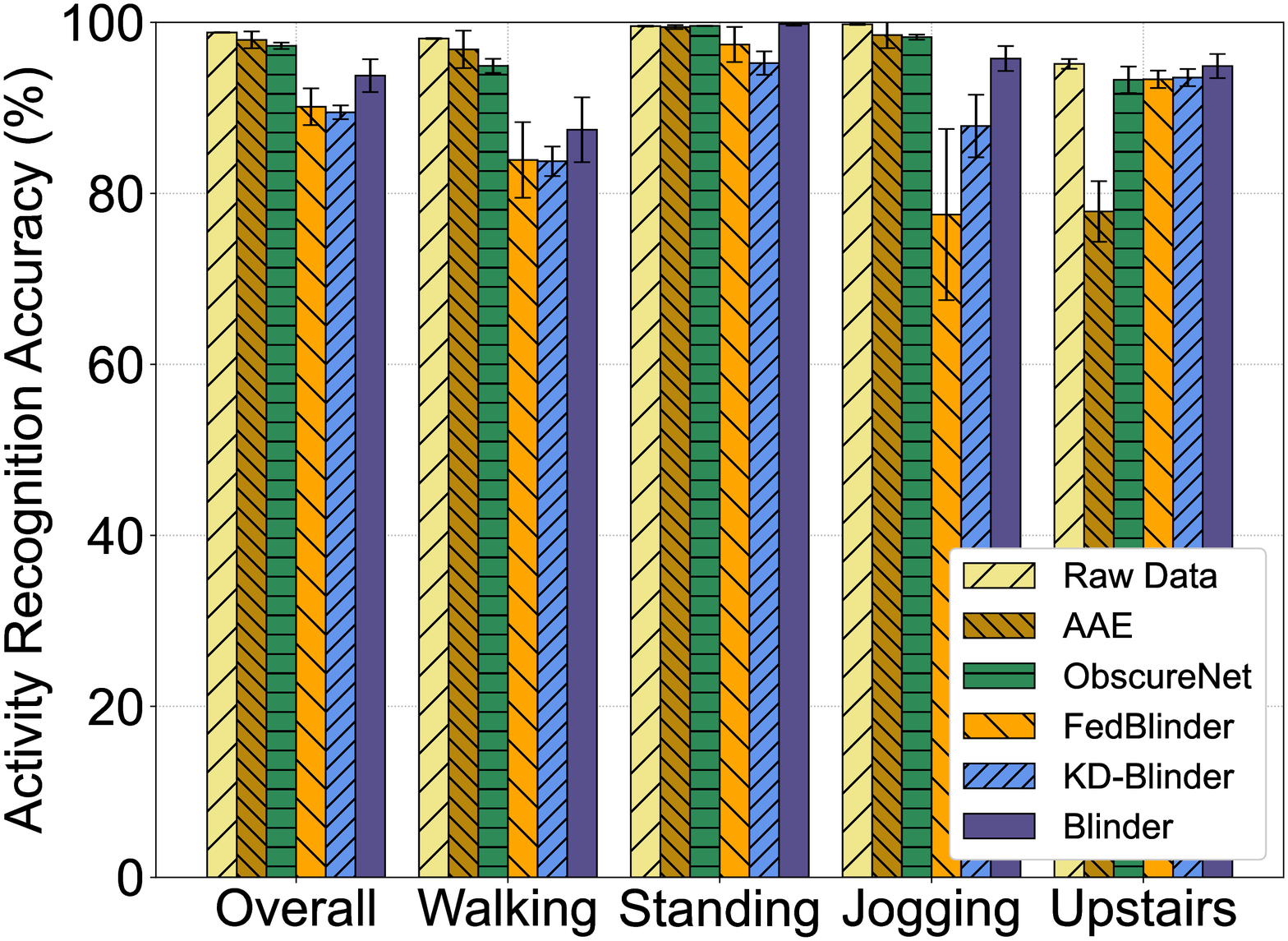}
         \caption{Gender anonymization}
         \label{fig:utility_random_mobi_gen}
     \end{subfigure}
     \hfill
     \begin{subfigure}[b]{0.49\linewidth}
         \centering
         \includegraphics[width=\linewidth]{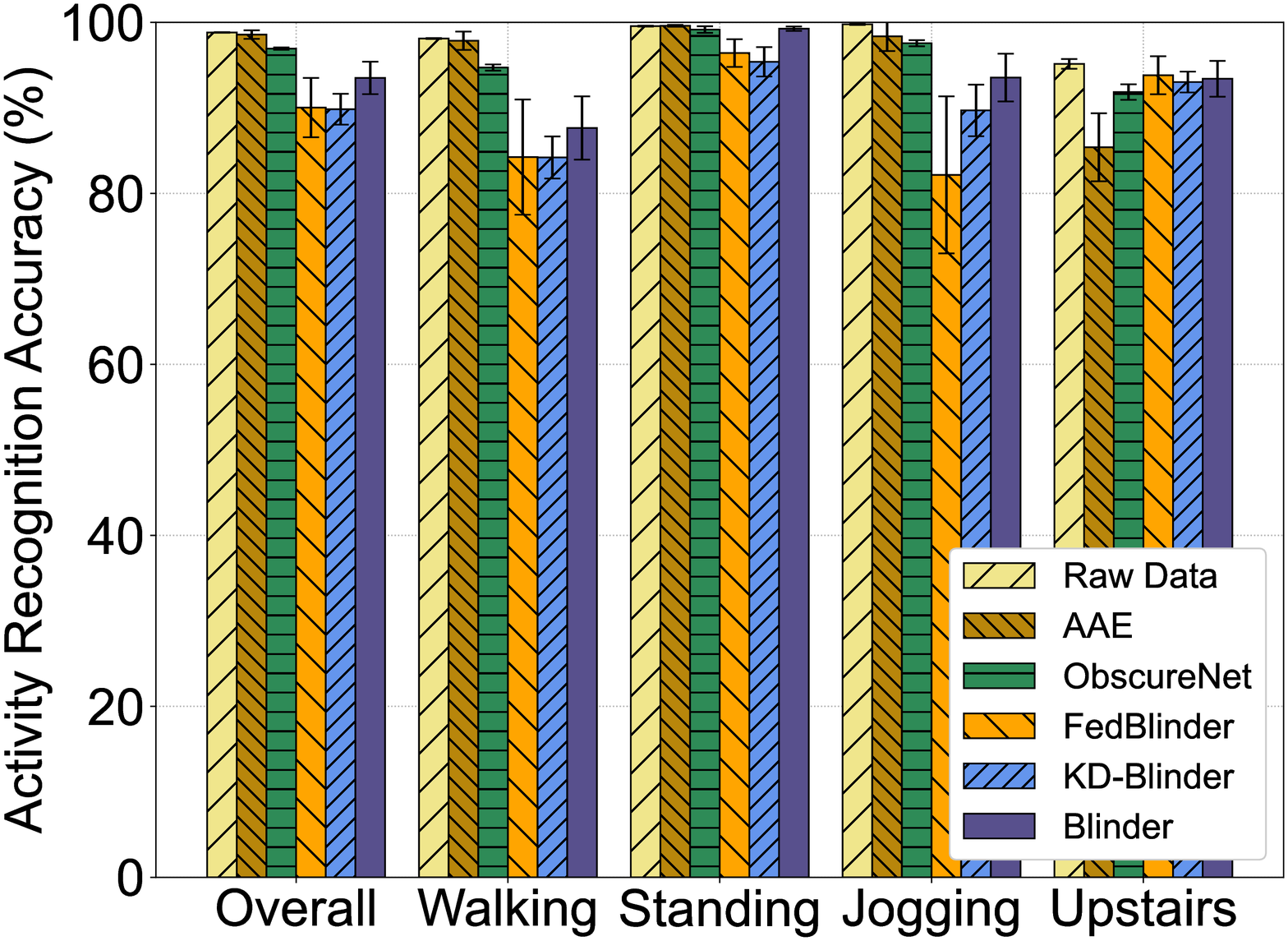}
         \caption{Weight anonymization}
         \label{fig:utility_random_mobi_weight}
     \end{subfigure}
    \caption{Activity recognition accuracy on MobiAct for gender and weight anonymization.}
    \label{fig:utility_random_mobi}
\end{figure}

\begin{figure}[t]
    \centering
    \begin{subfigure}[b]{0.49\linewidth}
         \centering
         \includegraphics[width=\linewidth]{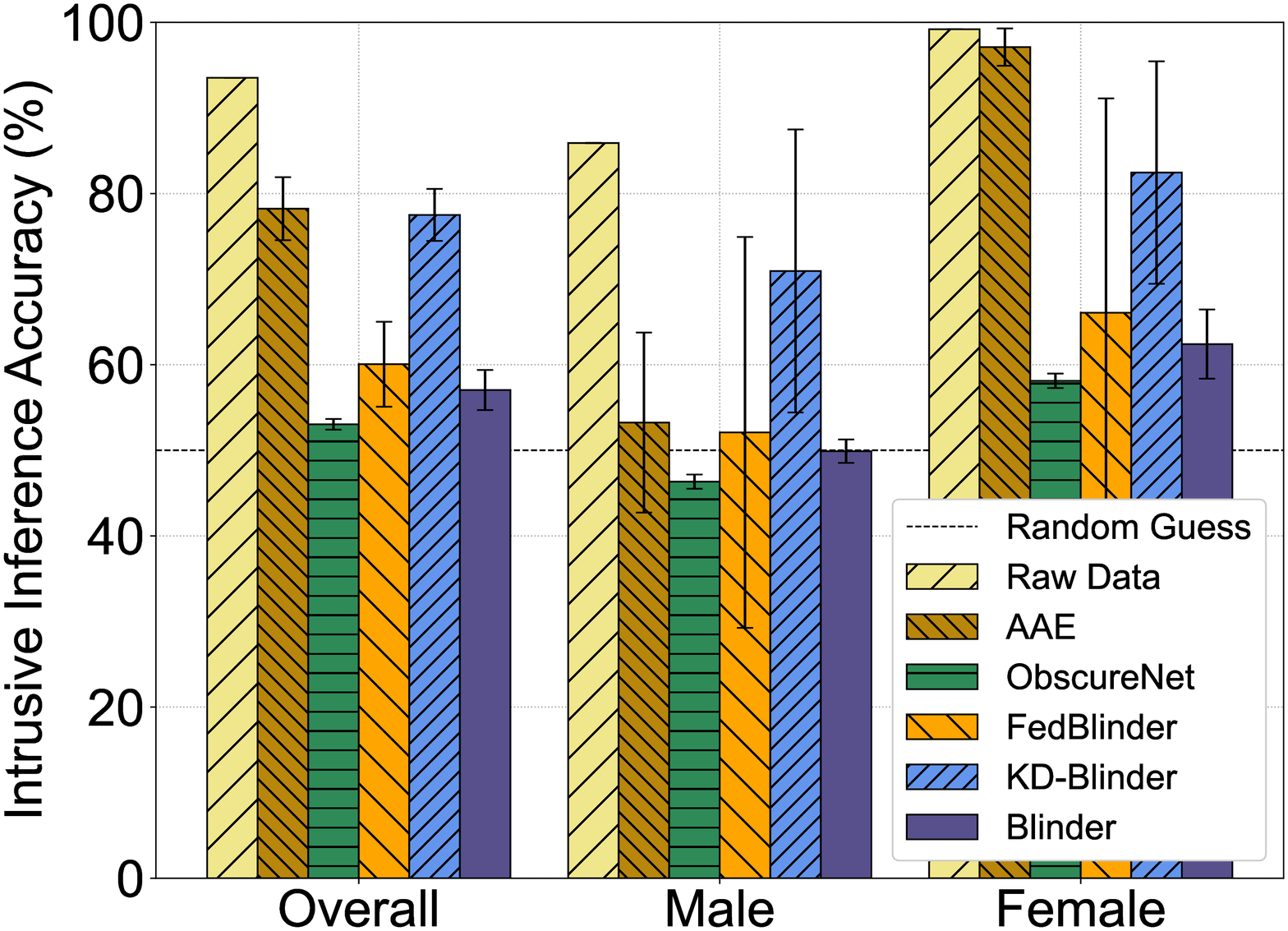}
         \caption{Intrusive inference accuracy}
         \label{fig:privacy_random_motion_gen}
     \end{subfigure}
     \hfill
     \begin{subfigure}[b]{0.49\linewidth}
         \centering
         \includegraphics[width=\linewidth]{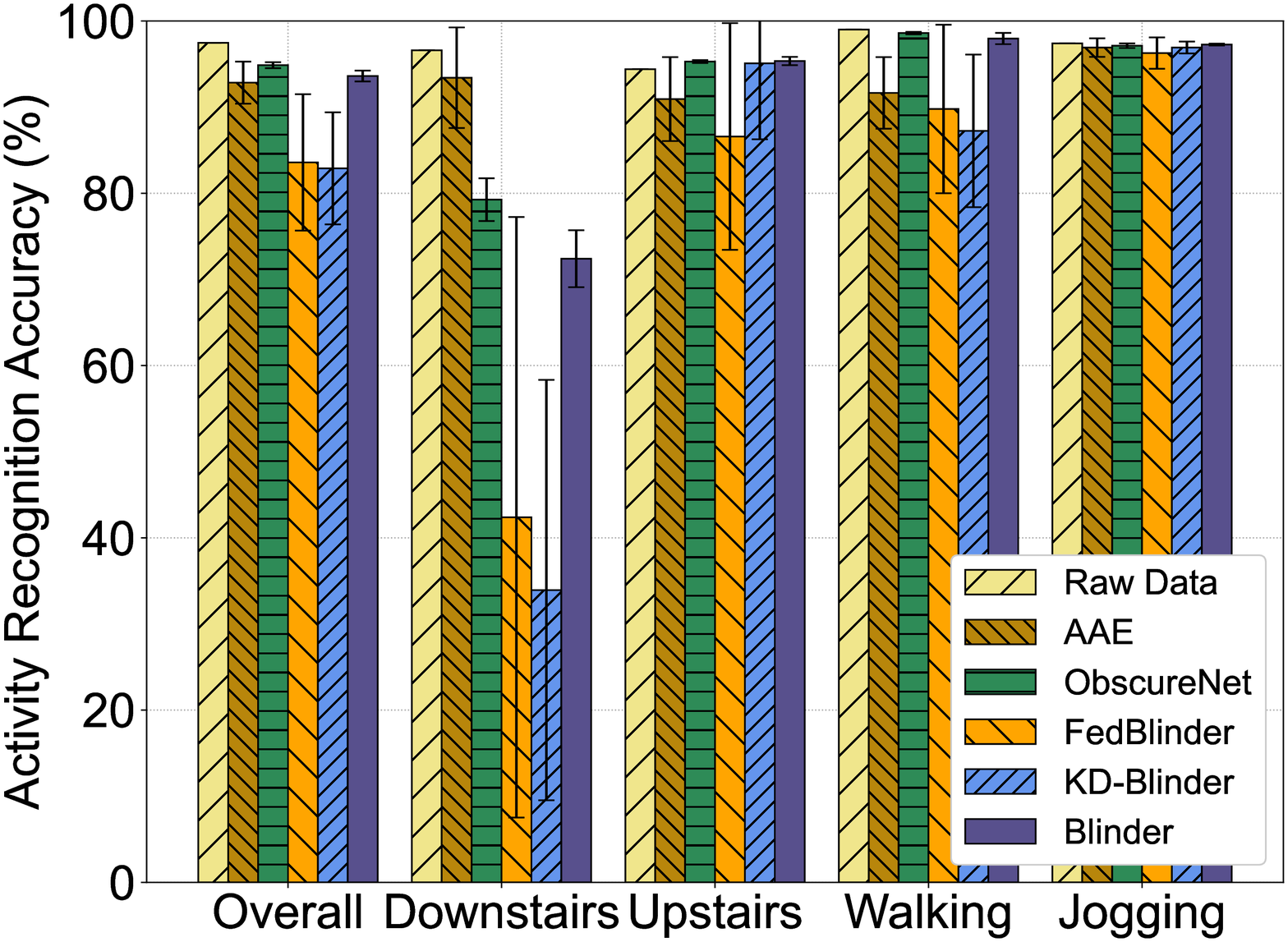}
         \caption{Activity recognition accuracy}
         \label{fig:utility_random_motion_gen}
     \end{subfigure}
    \caption{Intrusive inference and activity recognition accuracy on MotionSense for gender anonymization.}
    \vspace{-3mm}
    \label{fig:privacy_utility_random_motion}
\end{figure}

Figure~\ref{fig:privacy_random_motion_gen} compares Blinder and the baselines 
in terms of their privacy-preserving performance (i.e., obscuring gender) on MotionSense. 
In general, the performance of the five anonymization models follows the same pattern as in MobiAct, 
with ObscureNet having the best performance and 
    AAE having the worst performance. 
    The average intrusive inference accuracy and F1 score under ObscureNet (AAE) are $53.04\%$ and $52.21\%$ ($78.23\%$ and $75.39\%$), respectively.
    In particular, we find AAE can only effectively anonymize the male gender, 
    whereas the intrusive inference model can accurately predict the anonymized data with the female gender.
    FedBlinder performs better than AAE and slightly worse than Blinder, 
    with average intrusive inference accuracy and F1 score of $60.06\%$ and $56.8\%$, respectively.
Similar to MobiAct, FedBlinder suffers from the overfitting issue.
KD-Blinder performs significantly worse than FedBlinder 
but shows better consistency over the 10 runs with an average accuracy (F1 score) of $77.50\%$ ($76.31\%$).
The overall intrusive inference accuracy and F1 score under Blinder are $57.04\%$ and $56.15\%$, respectively, underperforming ObscureNet by only $4\%$. 
Despite the minor decline in the anonymization performance of Blinder compared to ObscureNet, 
which can be attributed to its distributed training, 
Blinder still outperforms FedBlinder and KD-Blinder, and reduces the privacy loss by $36.48\%$ 
compared to the case that the sensor data is not anonymized.

\subsection{Impact on Data Utility}
\label{subsec:eval_utility}
Figure~\ref{fig:utility_random_mobi} compares the utility of data for the desired inference task, i.e., HAR,
using different models to anonymize the MobiAct dataset. %
    It can be seen that the two centralized baselines yield the highest data utility. 
    Specifically, AAE attains average activity recognition accuracy of $97.95\%$ ($98.57\%$) 
    and F1 score of $86.69\%$ ($89.55\%$) in the gender (weight) anonymization task, 
    outperforming ObscureNet in terms of activity recognition accuracy by $0.70\%$ ($1.64\%$) and F1 score by $1.53\%$ ($5.49\%$). 
    When looking at each individual activity class, 
    we find that ObscureNet achieves more consistent data utility, 
    whereas AAE sacrifices the data utility for the upstairs activity.
    The data anonymized by FedBlinder and KD-Blinder have nearly the same utility, 
    which is the lowest among all models in both anonymization tasks.
    FedBlinder achieves average activity recognition accuracy of $90.13\%$ ($90.03\%$) in the gender (weight) anonymization task, outperforming KD-Blinder by $0.65\%$ ($0.19\%$).
    Looking at individual activities, the classification accuracy of `jogging' under FedBlinder is particularly low, 
    i.e., more than $20\%$ ($15\%$) below the classification accuracy of the same activity under ObscureNet in the gender (weight) anonymization task.
    With the help of the inter-class balancing techniques, KD-Blinder improves the classification accuracy of `jogging' by $12.86\%$ ($7.54\%$) compared to FedBlinder when anonymizing gender (weight).
    Thanks to the meta-learning-based personalized federated learning
    and rebalancing of the public attribute distribution, 
    Blinder outperforms both FedBlinder and KD-Blinder in both anonymization tasks. 
The average overall activity recognition accuracy on data anonymized by Blinder is $93.77\%$ ($92.69\%$) in the gender (weight) anonymization task, only $3.49\%$ ($4.24\%$) less than the ObscureNet baseline.
Looking at the breakdown per activity, 
the activity recognition accuracy under Blinder is always higher than $85\%$.
For the walking activity that has the lowest accuracy, 
we find that most of the misclassified samples are confused with climbing up the stairs.
This is due to their similar movement patterns in the horizontal plane %
with the main difference being in the vertical plane. %
This result is encouraging given that Blinder is trained in a distributed fashion. %

In Figure~\ref{fig:utility_random_motion_gen}, 
we report the utility of data for the HAR task on MotionSense when anonymizing gender. 
ObscureNet achieves %
the highest utility, %
resulting in average activity recognition accuracy and F1 score of $94.87\%$ and $92.93\%$, respectively. 
Blinder (FedBlinder) achieves an average activity recognition accuracy of $93.61\%$ ($83.58\%$),
which is $1.26\%$ ($11.29\%$) lower than ObscureNet, and an F1 score of $91.18\%$ ($77.11\%$).
    Despite being trained in a centralized manner, AAE's performance is on par with Blinder, 
    with a slightly lower activity recognition accuracy of $92.84\%$, 
    but a higher F1 score of $91.66\%$. 
    This is because AAE can provide more consistent data utility across all activities, 
    especially for the challenging activity of going downstairs.
    The data anonymized by both FedBlinder and KD-Blinder have significantly lower utility, with KD-Blinder yielding a $0.69\%$ ($0.99\%$) lower average activity recognition accuracy (F1 score) than FedBlinder. This is mainly due to the difficulty of differentiating between walking up and down the stairs.
Similarly, we observe that both Blinder and ObscureNet perform the worst on walking down the stairs, 
with an average accuracy of $72.4\%$ and $79.26\%$, respectively. 
We believe this is because of the data quality issues in MotionSense. 
Similar observations are made in~\cite{malekzadeh2019mobile} 
where the latent representations of walking up and down the stairs are overlapping,
and in~\cite{hajihassnai2021obscurenet} where walking down the stairs had the lowest activity recognition accuracy.
    This data quality issue could render the inter-class balancing techniques used in KD-Blinder less effective as they might further deteriorate the quality of data, 
    resulting in even lower data utility.
    Blinder, however, generates high-quality data samples 
    using the SMOTE-based data synthesis technique.
Note Blinder trains a single model to protect data for all activities whereas ObscureNet protects each activity using a separate model. Hence, there is a higher probability for Blinder to confuse walking down the stairs
with activities that have similar movement patterns.

\begin{figure}[t]
    \centering
    \begin{subfigure}[b]{0.49\linewidth}
         \centering
         \includegraphics[width=\linewidth]{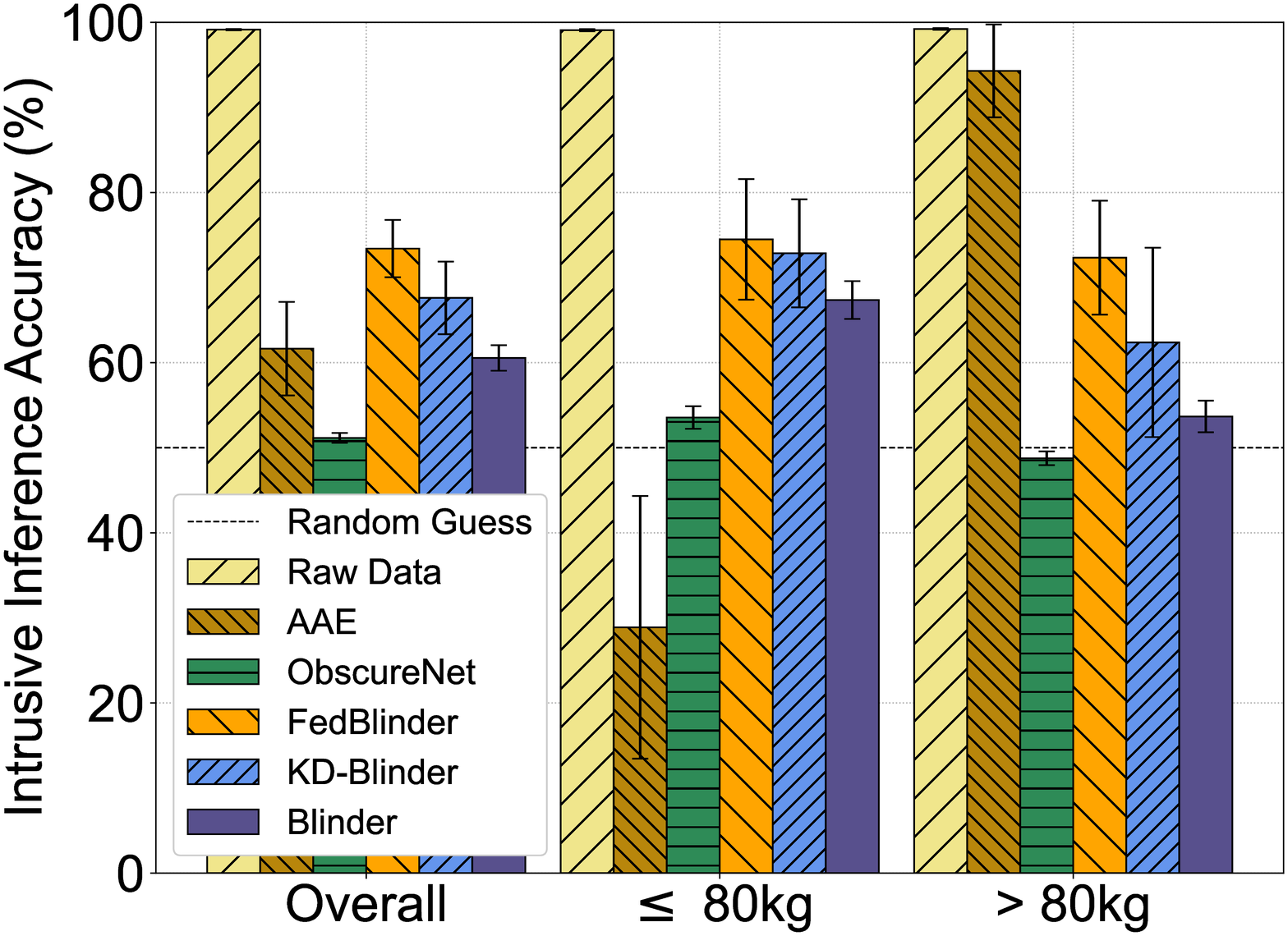}
         \caption{Weight anonymization}
         \label{fig:privacy_random_wifi_weight}
     \end{subfigure}
     \hfill
     \begin{subfigure}[b]{0.49\linewidth}
         \centering
         \includegraphics[width=\linewidth]{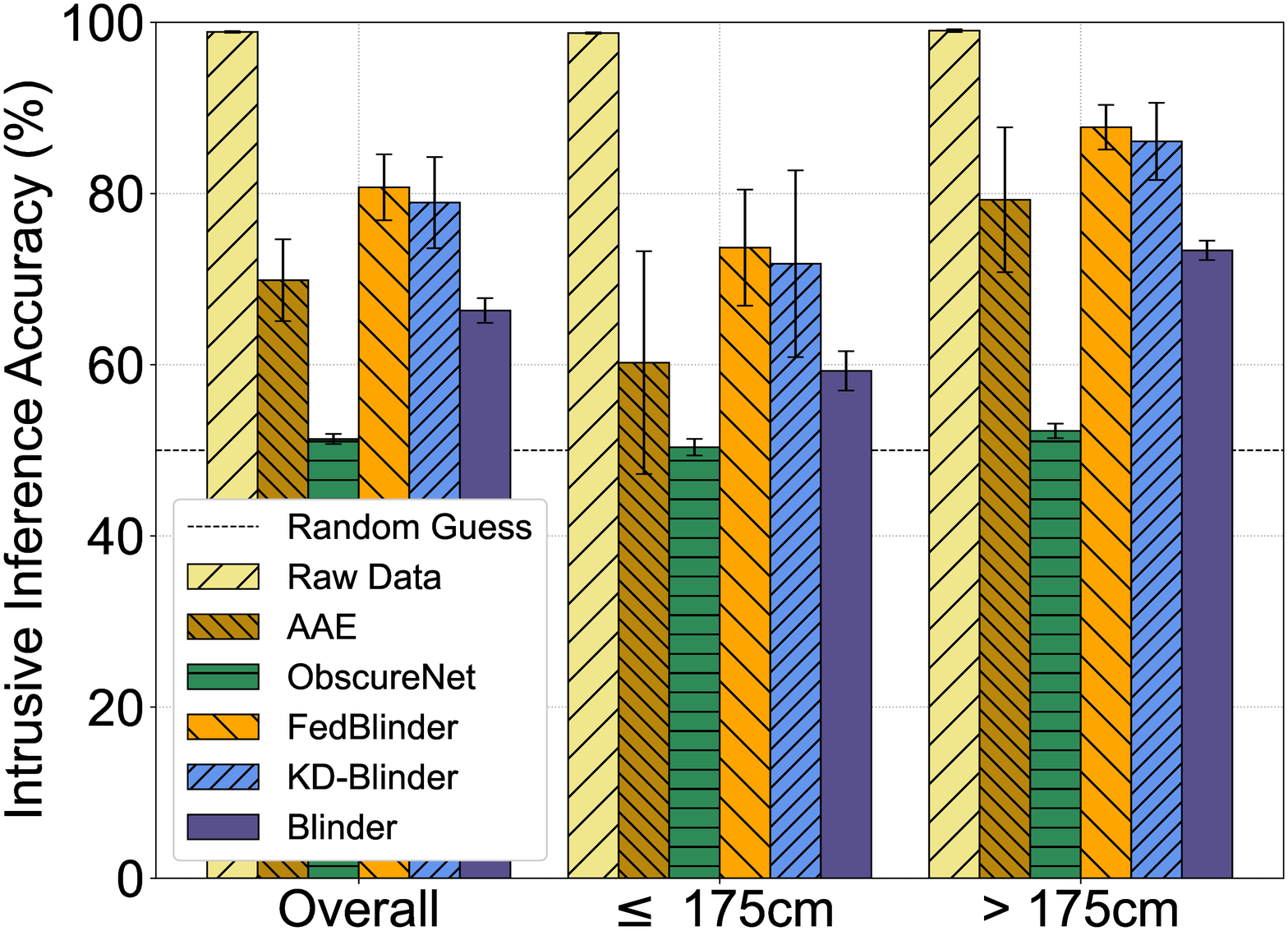}
         \caption{Height anonymization}
         \label{fig:privacy_random_wifi_height}
     \end{subfigure}
    \caption{Intrusive inference accuracy on the Wi-Fi HAR dataset for weight and height anonymization.}
    \vspace{-3mm}
    \label{fig:privacy_random_wifi}
\end{figure}

\subsection{Data Anonymization on Wi-Fi Signals}
\begin{figure}[t]
    \centering
    \begin{subfigure}[b]{0.49\linewidth}
         \centering
         \includegraphics[width=\linewidth]{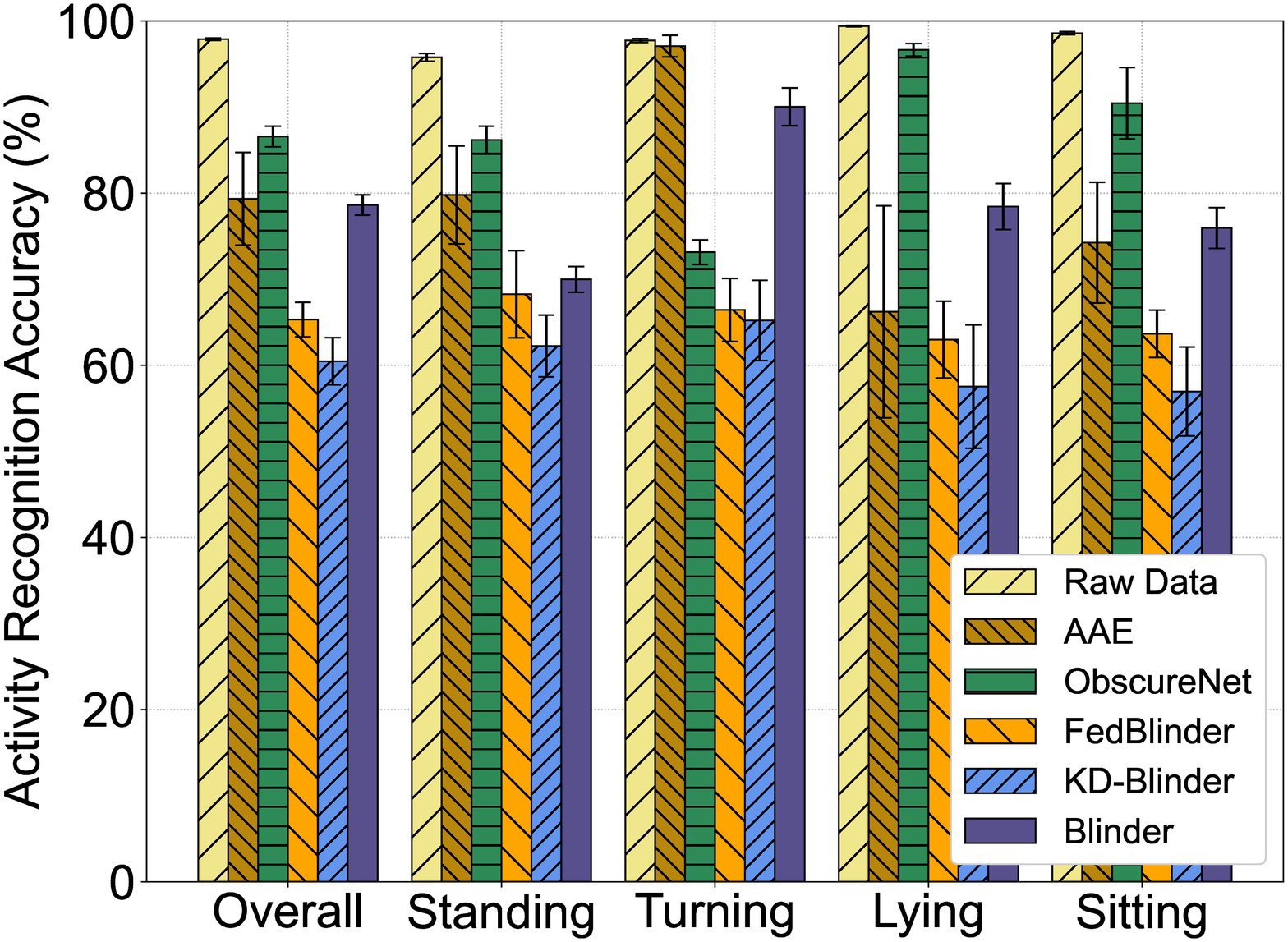}
         \caption{Weight anonymization}
         \label{fig:utility_random_wifi_weight}
     \end{subfigure}
     \hfill
     \begin{subfigure}[b]{0.49\linewidth}
         \centering
         \includegraphics[width=\linewidth]{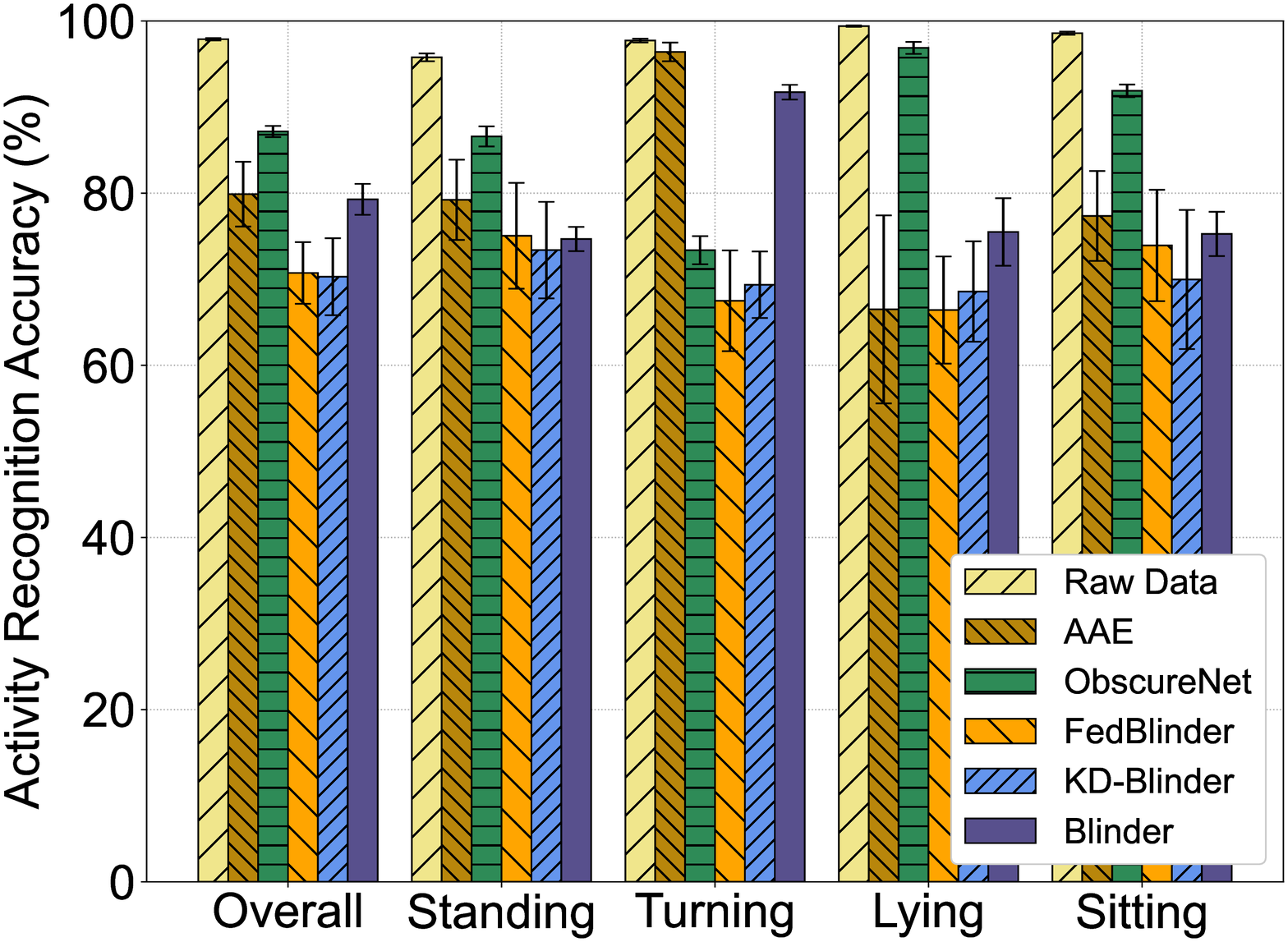}
         \caption{Height anonymization}
         \label{fig:utility_random_wifi_height}
     \end{subfigure}
    \caption{Activity recognition accuracy on the Wi-Fi HAR dataset for weight and height anonymization.}
    \label{fig:utility_random_wifi}
    \vspace{-4mm}
\end{figure}

    In addition to the two IMU sensor datasets, we further study Blinder's data anonymization capability on the Wi-Fi HAR dataset. 
    We illustrate the intrusive inference accuracy in Figure~\ref{fig:privacy_random_wifi} and activity recognition accuracy in Figure~\ref{fig:utility_random_wifi} for weight and height anonymization.

    When anonymizing the binary weight group attribute, 
    ObscureNet exhibits the best privacy-preserving capability, 
    achieving average accuracy (F1 score) of $51.15\%$ ($51.12\%$). 
        The overall intrusive inference accuracy of Blinder is marginally better (closer to the random guessing level) than 
        the centralized AAE baseline, %
        yet it still underperforms ObscureNet by $9.40\%$. 
        The distributed FedBlinder and KD-Blinder baselines 
        underperform Blinder by $12.86\%$ and $17.33\%$, respectively. 
    We find that although the overall intrusive inference accuracy of AAE is similar to Blinder, 
    AAE is only effective in protecting the weight group of under 80kg 
    and does not do a great job for the other weight group.
    Similar observations can be made when anonymizing the height attribute as shown in Figure~\ref{fig:privacy_random_wifi_height}. 
    The overall intrusive inference accuracy of Blinder is $66.33\%$, outperforming the centralized AAE baseline by $3.54\%$ with more consistent performance across $10$ trials. 
    Similar to anonymizing the two IMU datasets 
    and the weight attribute in the Wi-Fi HAR dataset, 
        FedBlinder and KD-Blinder have the worse performance among all anonymization models in this case.

    Next, we look at the data utility on the Wi-Fi HAR dataset. 
    When performing weight anonymization, the data anonymized by ObscureNet retains the highest utility, attaining average accuracy (F1 score) of $86.58\%$ ($86.41\%$).
    Blinder and AAE yield a similar activity recognition accuracy (F1 score) of $78.62\%$ ($78.51\%$) and $79.34\%$ ($79.22\%$), respectively, although AAE cannot effectively protect the privacy of both weight groups.
        KD-Blinder performs the worst with respect to data utility, 
        achieving activity recognition accuracy of $\sim60\%$, 
        which is about $5\%$ lower than FedBlinder.
    When anonymizing the height attribute, we find that all models perform similarly compared to anonymizing the weight attribute. 
    Specifically, ObscureNet yields the highest average accuracy (F1 score) of $87.16\%$ ($86.99\%$).
    Blinder achieves an average accuracy (F1 score) of $79.28\%$ ($79.16\%$), which is similar to AAE that shows an average accuracy of $79.88\%$ and outperforms FedBlinder and KD-Blinder. 
    The results show that Blinder can be used to anonymize the RF sensing modality 
    without sacrificing the HAR accuracy. 
    Since Blinder can compete with state-of-the-art data anonymization methods 
    that are trained on centralized data,
    we believe it is a promising anonymization technique as it provides 
    strong privacy protection in the entire data consumption life cycle.
    
\begin{figure}[t]
\includegraphics[width=0.6\linewidth]{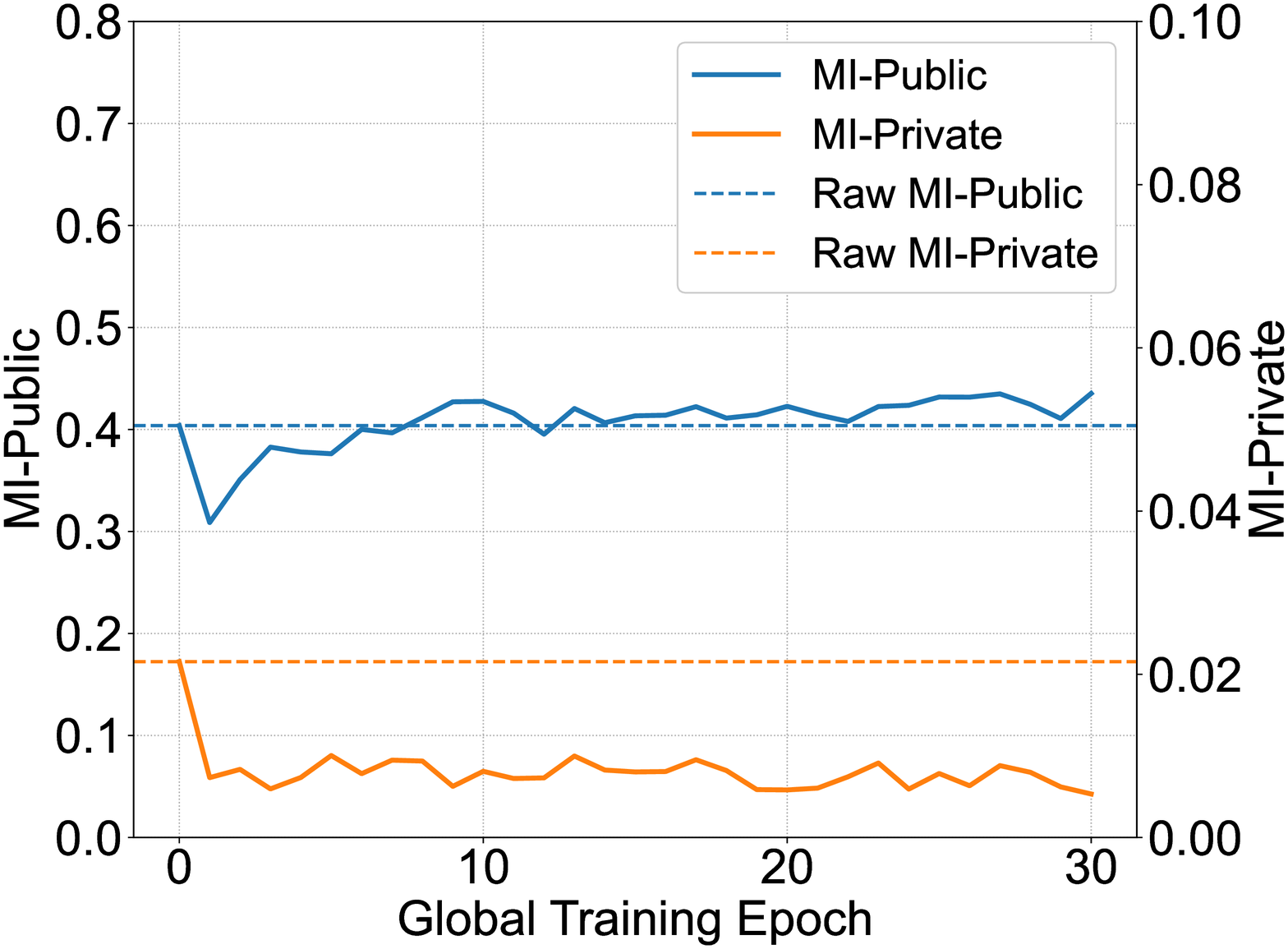}
\vspace{-2mm}
\caption{
    Mutual information between the low-dimensional representation of sensor data and 
    public/private attributes for the gender anonymization task on MobiAct.
} 
    \label{fig:mutual_info}
    \vspace{-2mm}
\end{figure}

\subsection{Information-theoretic Validation of Blinder}
\label{subsec:external_validation}
So far we have used deep CNN-based intrusive and desired inference models 
to show that an acceptable privacy-utility trade-off can be achieved using Blinder.
We now take an information-theoretic approach to demonstrate Blinder's privacy-preserving capability,
regardless of the inference models.
We use Principal Component Analysis (PCA)~\cite{tipping1999mixtures} 
to project each data segment into a low-dimensional space, 
thereby extracting the most significant $n$ feature components.
This is done for the raw sensor data and the anonymized sensor data. 
We then use Mutual Information (MI)~\cite{kozachenko1987sample}  
to measure the mutual dependency between this representation and the public or private attribute. 
Hence, higher MI indicates a stronger dependency between sensor data and the corresponding attribute.

Figure~\ref{fig:mutual_info} shows the average MI between the principal components 
extracted from the anonymized sensor data and their corresponding public/private attributes
as we increase the number of global training epochs.
To obtain this result, we performed stochastic anonymization on MobiAct, 
assuming activity is the public attribute and gender is the private attribute. 
We let $n=25$ so that it is equal to Blinder's latent space dimension.
The dashed (horizontal) line shows MI between the raw sensor data and their corresponding public/private attributes. 
We observe that Blinder is able to reduce the dependency between the anonymized data and the private attribute 
by around $75\%$ at the end of the first global training epoch. 
However, the average MI between sensor data and the public attribute also decreases 
at the early stage of training, indicating the data utility is initially sacrificed to obscure the privacy attribute.
As Blinder converges, the average MI between sensor data and the public attribute returns %
to its initial level.
Meanwhile, the average MI between sensor data and the private attribute remains at a low level.

\subsection{Obscuring Multiple Private Attributes}\label{sec:multiple}
We consider the practical yet challenging case 
when a user attempts to protect multiple private attributes simultaneously. 
Using the MobiAct dataset as an example, 
we set activity as the public attribute and both gender and weight groups as private attributes.
In this case, two discriminators are used in Blinder to ensure 
that gender and weight group cannot be inferred from the latent representation.
Moreover, we concatenate the latent representation with both private attributes and the public attribute
before it is sent to Blinder's decoder.
We repeat the experiments $10$ times and directly compare Blinder's performance 
with ObscureNet's performance as reported in~\cite{hajihassnai2021obscurenet}. 
Specifically, the average activity recognition accuracy (F1 score) under Blinder is $93.64\%$ ($78.17\%$),
which is only $2.07\%$ lower than ObscureNet's performance. 
For anonymizing the gender (weight group) attribute, the average intrusive inference accuracy under Blinder is 
$58.36\%$ ($55.05\%$), which is around $6.2\%$ ($4.79\%$) higher than ObscureNet's performance. 
To put it in another context, when anonymizing both gender and weight, 
Blinder can reduce the accuracy of gender and weight-group identification by $39.15\%$ and $36.62\%$ respectively,
compared to the case where data is not anonymized.
This suggests Blinder can protect multiple private attributes 
by incorporating additional discriminator modules as needed.

\begin{figure}[t]
    \centering
    \begin{subfigure}[b]{0.49\linewidth}
         \centering
         \includegraphics[width=\linewidth]{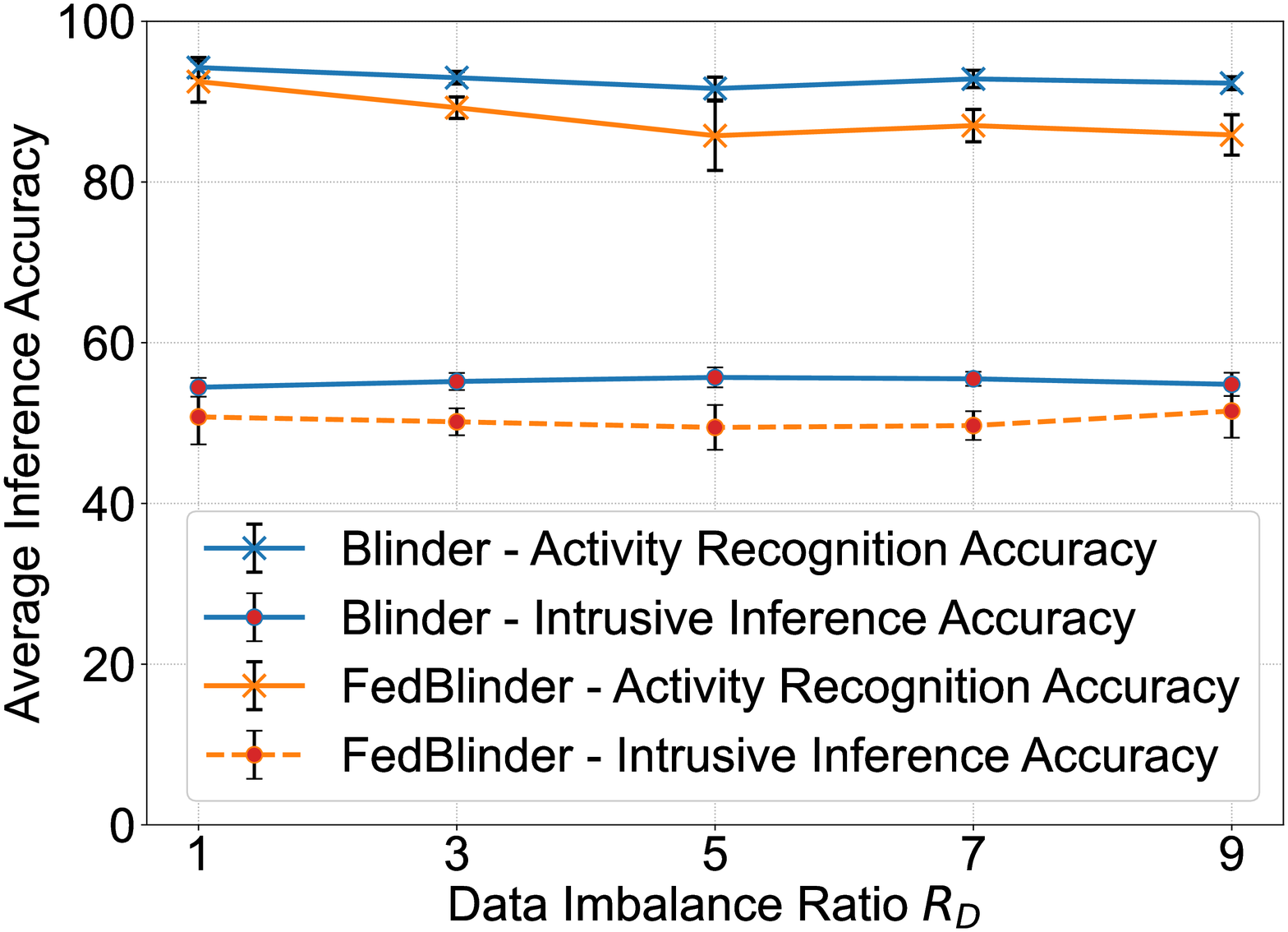}
         \vspace{-3mm}
         \caption{Gender anonymization}
         \label{fig:hetero_data_gen}
     \end{subfigure}
     \hfill
     \begin{subfigure}[b]{0.49\linewidth}
         \centering
         \includegraphics[width=\linewidth]{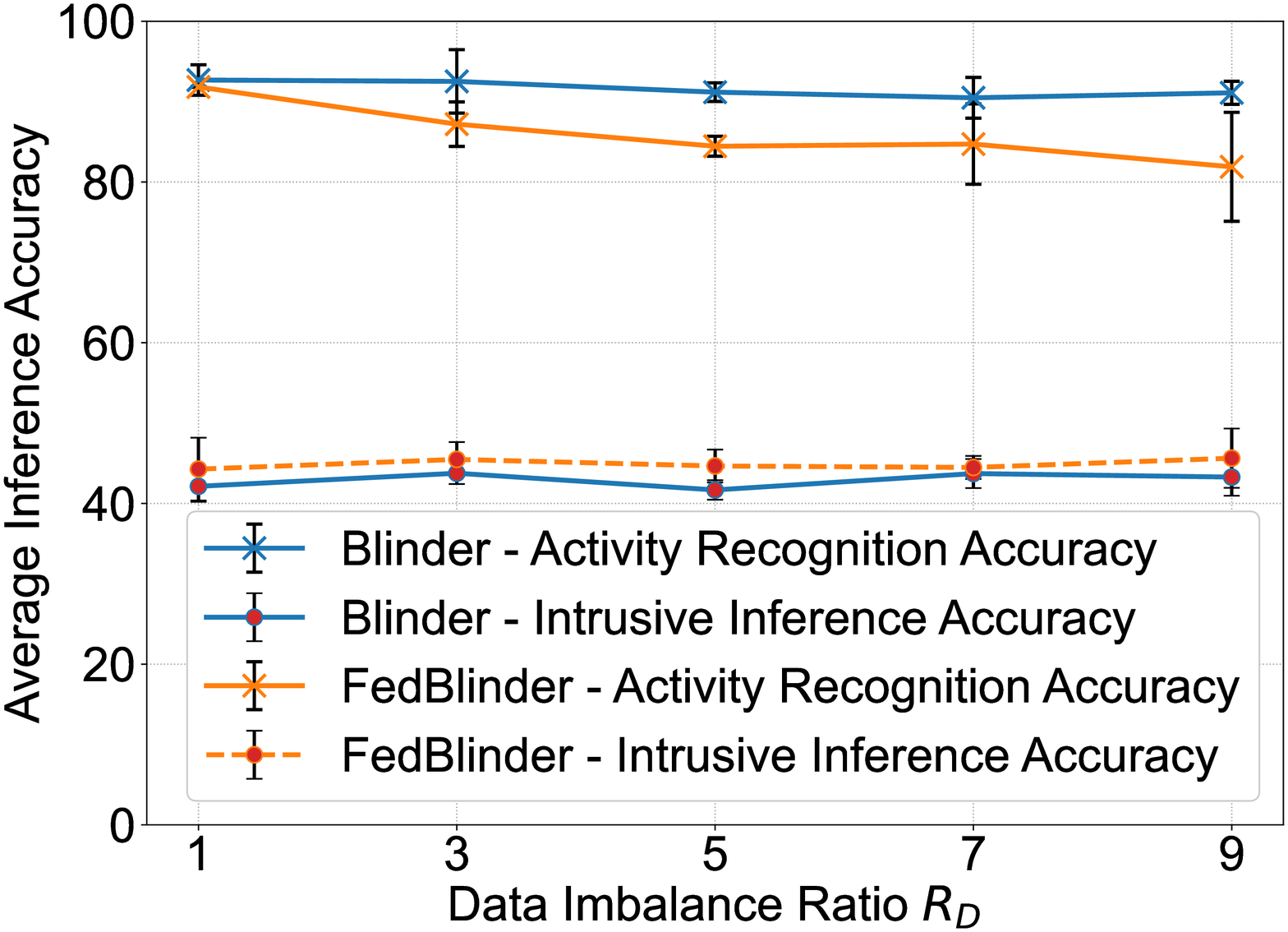}
        \vspace{-3mm}
         \caption{Weight anonymization}
         \label{fig:hetero_data_weight}
     \end{subfigure}
     \vspace{-2mm}
    \caption{Impact of heterogeneous training data on Blinder and FedBlinder in terms of activity recognition and intrusive inference on MobiAct for gender and weight anonymization.}
    \vspace{-4mm}
    \label{fig:hetero_data}
\end{figure}

\subsection{Addressing Heterogeneity}
We compare the generalizability of the anonymization models 
when the clients participating in model training have non-i.i.d. data distributions.
Note that ObscureNet bypasses the non-i.i.d. problem 
by using attribute-specific models, which are costly to train in practice. 
Hence, we exclude ObscureNet from this evaluation.
We vary $R_D$ from $1$ to $9$, i.e., from balanced i.i.d. datasets 
to the case where each user's majority public attribute class
has $9\times$ more samples than the respective minority class.   
We compare the model's anonymization capability in gender and weight anonymization tasks on MobiAct,
and show the result in Figure~\ref{fig:hetero_data}.
It can be seen that the increased skewness in public attribute classes 
does not have a noticeable impact on the privacy-preserving capability of the anonymization models;
both Blinder and FedBlinder have stable performance.
However, we notice that data utility declines substantially under FedBlinder
with more skewed local public attribute distributions;
the overall activity recognition accuracy in the gender (weight) anonymization task 
goes down from $92.49\%$ ($91.83\%$) when $R_D=1$ to $85.86\%$ ($81.89\%$) when $R_D=9$.
This implies that FedBlinder is more susceptible to imbalanced, non-i.i.d. public attribute distributions. 
As we discussed in Section~\ref{subsec:eval_utility}, 
FedBlinder achieves activity recognition accuracy of around ${90\%}$ on MobiAct, 
which contains even more imbalanced public attribute classes ($R_D\approx40$). 
This is because the original MobiAct dataset only contains imbalanced public attribute classes, 
but the class distributions are almost identical among the users.

Blinder, however, is more robust to the imbalanced, non-i.i.d. data 
and causes less than $2\%$ decline in data utility
in both gender and weight anonymization tasks when increasing $R_D$ from $1$ to $9$.
This shows that even without the public attribute rebalancing technique, 
Blinder can tackle the non-i.i.d. data problem, where the public attribute classes are heavily imbalanced.

\begin{figure}[tb]
    \centering
    \begin{minipage}[t]{.49\linewidth}
        \centering
        \includegraphics[width=\linewidth]{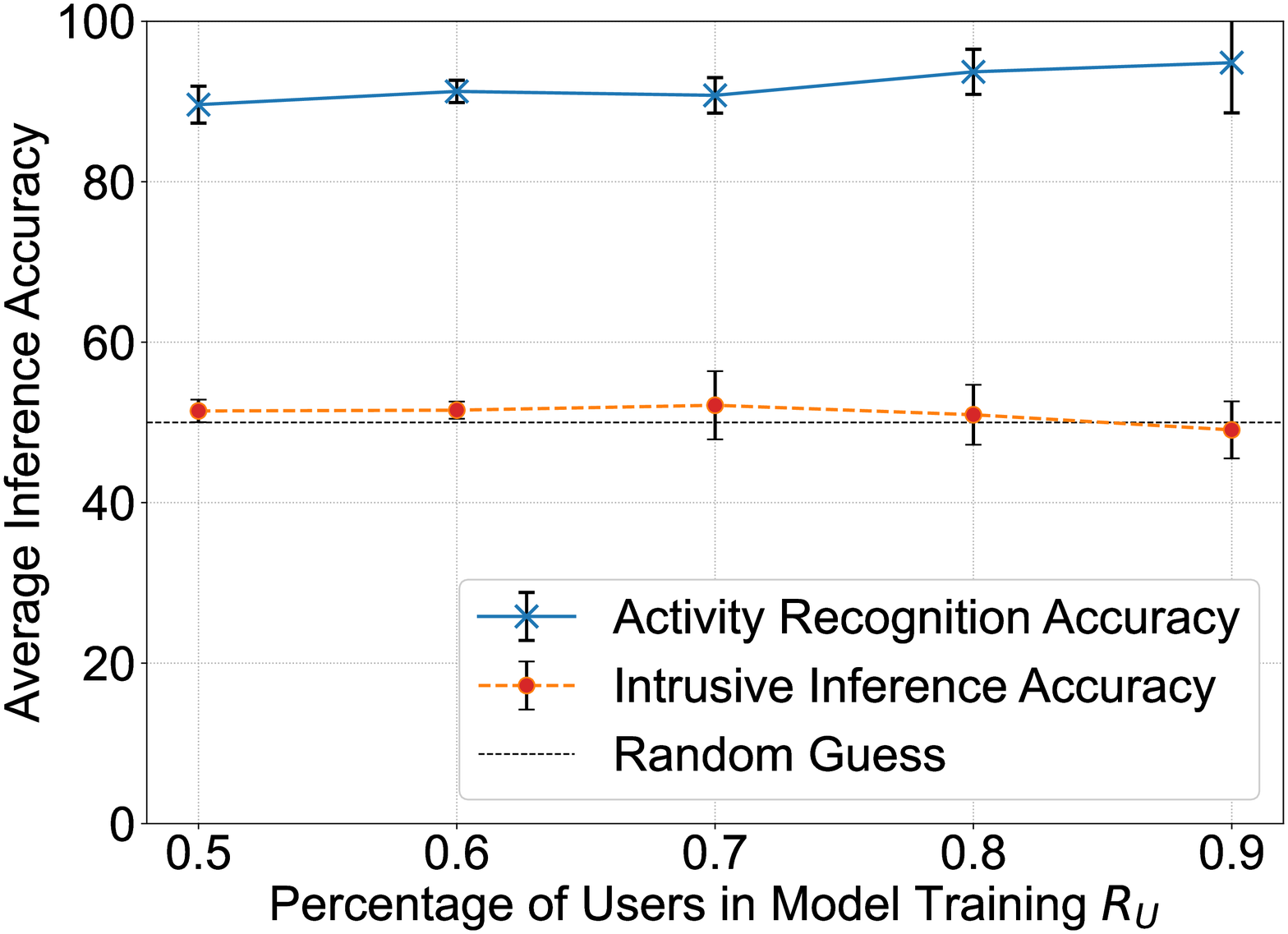}
        \vspace{-5mm}
        \caption{Generalization to unseen users on MobiAct dataset for gender anonymization. 
}
        \label{fig:hetero_users}
    \end{minipage}%
    \hfill
    \begin{minipage}[t]{0.49\linewidth}
        \centering
        \includegraphics[width=\linewidth]{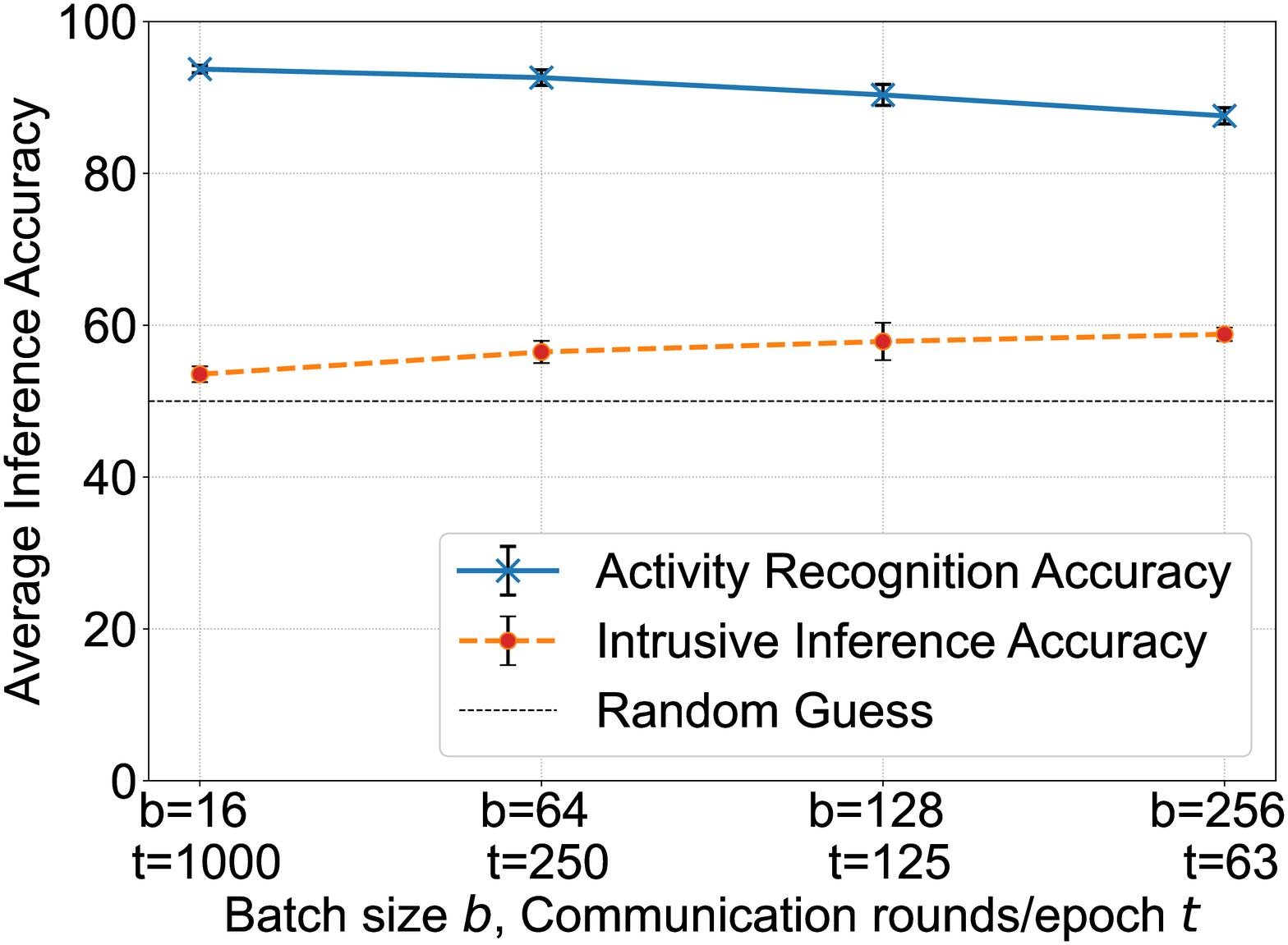}
        \vspace{-5mm}
        \caption{Communication cost and anonymization performance trade-off on MobiAct.}
        \label{fig:comm_mean_accu}
        \vspace{-2mm}
    \end{minipage}
\end{figure}

\subsection{Generalizing and Adapting to New Users}
\label{sec:adaptation}
We explore whether Blinder generalizes to users who have never contributed to model training. 
Since FedBlinder shows variable performance 
for different activities and private attributes,
here we only focus on evaluating Blinder.
Figure~\ref{fig:hetero_users} shows the accuracy of desired and intrusive inferences
averaged over $5$ trials on MobiAct, when Blinder is trained to perform gender anonymization.
Error bars indicate the standard deviation.
We vary $R_U$ from $0.5$ to $0.9$ with increments of size $0.1$. 
We do not consider $R_U < 0.5$ since Blinder's anonymization performance varies widely
across multiple runs due to the small number of clients that are being sampled. %
Results show that Blinder can effectively learn to anonymize gender attribute under various $R_U$ values. 
It achieves the best privacy-preserving performance when 32 users (about $90\%$ of all users)
collaboratively train the model, showing an average intrusive inference accuracy of $49.7\%$.
When Blinder is trained collaboratively by 18 users only ($50\%$ of the users), 
the intrusive inference accuracy increases slightly by $1.6\%$,
which is still close to the accuracy of random guessing.

Turning our attention to data utility, 
lowering the participation rate in model training to $R_U=0.5$ has a moderate impact on utility;
the average activity recognition accuracy drops from $94.84\%$ when $R_U=0.9$ 
to $89.61\%$ when $R_U=0.5$.
But, the lowest activity recognition accuracy among the 4 activities remains above $80\%$.
This confirms that Blinder generalizes well to unseen users.

Since Blinder is trained via the personalized federated learning algorithm,
we anticipate that it better generalizes to all users. 
Even if the pre-trained Blinder model does not perform optimally for the unseen users, 
one can personalize this model through local adaptation. 
We verify Blinder's personalization ability when $R_U=0.5$. 
We choose the $10$ unseen users who had the worst data utility 
after their data was anonymized using the pre-trained Blinder, 
with an average activity recognition accuracy of $84.47\%$ and intrusive inference accuracy of $50.61\%$. 
For each of these unseen users, we randomly sample a small portion ($<5\%$) of their local data 
and treat it as the adaptation set to personalize Blinder. 
The adaptation is performed on the user's device by running $80$ iterations of local training.
We observe the average data utility under the personalized Blinder 
increases by $5.1\%$ to $89.57\%$ across these $10$ users. %
Meanwhile, the intrusive (gender) inference 
accuracy remains under $\sim{51\%}$ after personalization. 
This confirms that Blinder's anonymization performance can be further enhanced for users
who did not participate in the model training through model personalization. %

\subsection{Trade-off between Communication Cost and Anonymization Performance}
We now study the trade-off between Blinder's anonymization performance and 
the communication overhead of the personalized federated learning algorithm used to train Blinder.
Without loss of generality, we assume that all users that participate in collaborative training 
of Blinder have the same number of data samples, denoted by $K$.
Each user divides its $K$ data samples into $t$ non-overlapping batches of data, each of size $b = s + q$, 
where $s$ is the size of the support set and $q$ is the size of the query set.
In each round of communication between the server and the selected client 
(Line~4 in Algorithm~\ref{alg:central_meta_obscurenet}),
the client uses one batch of data to update the parameters of 
its local anonymization model before it sends the gradients of the updated model to the server.
Hence, $t=\frac{K}{b}$ represents the total number of communication rounds per epoch.
Since the gradients that are sent to the server in each round have a fixed size,
the total communication overhead %
depends only on $t$ for a fixed number of epochs.
While a larger batch size reduces $t$ and the communication overhead accordingly, 
the client may need more local training steps 
(Line~16 to~21 is one training step in Algorithm~\ref{alg:central_meta_obscurenet}) 
for the model to converge, %
trading communication for local computation.
Thus, Blinder's anonymization performance depends on both the number of communication rounds 
and the number of local training steps.

\begin{figure}[tb]
    \centering
    \begin{subfigure}[t]{0.49\linewidth}
         \centering
         \includegraphics[width=\linewidth]{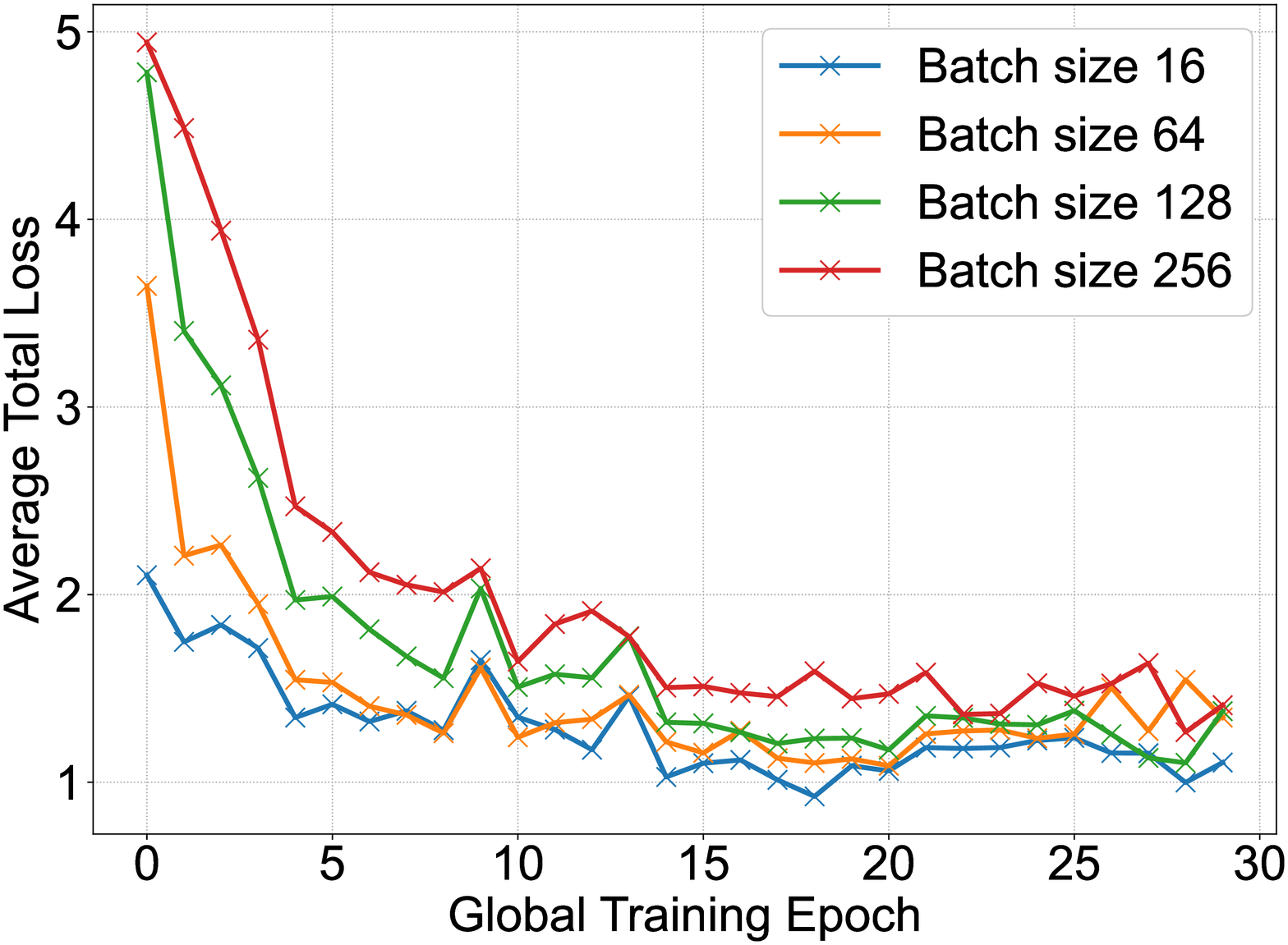}
         \caption{Impact of communication rounds on average total loss}
         \label{fig:comm_loss}
     \end{subfigure}
     \hfill
     \begin{subfigure}[t]{0.49\linewidth}
         \centering
         \includegraphics[width=\linewidth]{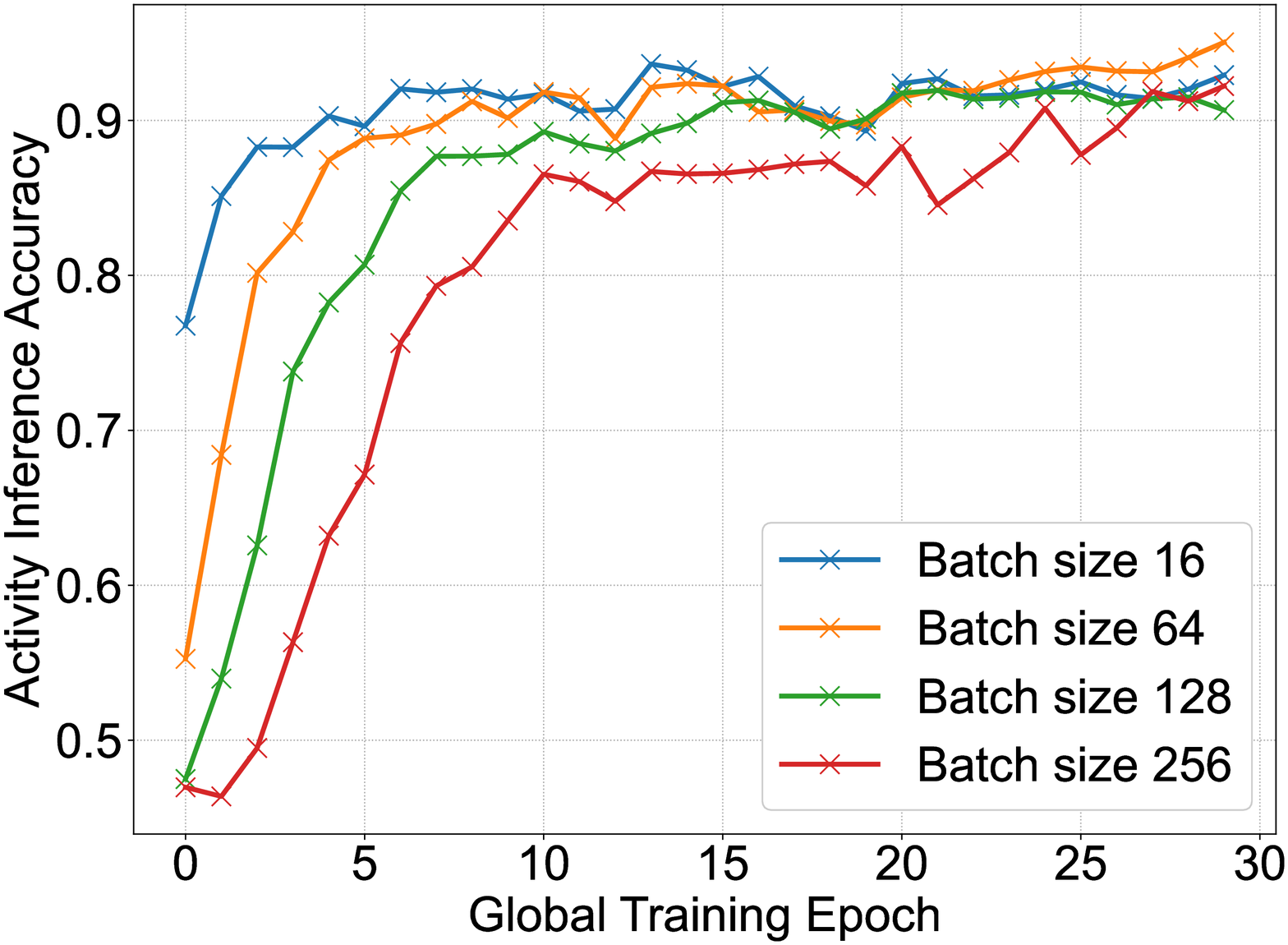}
         \caption{Impact of communication rounds on data utility}
         \label{fig:comm_accu}
     \end{subfigure}
    \caption{Blinder's gender anonymization performance for different batch sizes and global training epochs on MobiAct.}
    \label{fig:comm_tradeoff}
\end{figure}

We tweak the batch size ($b \propto \frac{1}{t}$) to navigate 
the trade-off between communication cost and anonymization performance.
So far in our experiments, we have used a batch size $b=s+q=16$ with $s=1, q=15$.
We increase $s$ and $q$ proportionally to obtain the following batch sizes: $64$, $128$, and $256$. 
To compensate for fewer communication rounds when using a larger batch size,
we proportionally increase the number of local training steps. 
Figure~\ref{fig:comm_loss} shows an example of Blinder's training loss 
for different batch sizes and communication costs when performing gender anonymization on MobiAct. 
It can be readily seen that using fewer communication rounds per epoch
slows down the convergence of Blinder when training for the same amount of global epochs.
As a result, the activity inference accuracy achieved by fewer communication rounds per epoch 
is reduced compared to the accuracy achieved by more communication rounds but fewer local training steps 
as shown in Figure~\ref{fig:comm_accu}.
We illustrate Blinder's anonymization performance over $5$ runs in Figure~\ref{fig:comm_mean_accu}.
Blinder performs better when it is trained with more communication rounds (accordingly fewer local training steps).
It achieves average activity recognition of $93.75\%$ 
and average intrusive inference accuracy of $53.55\%$ when $b=16, t=1000$,
Increasing the batch size to $256$ ($t = 63$), the activity recognition accuracy falls by around $6\%$
and the intrusive inference accuracy increases by $5\%$.
Indeed, the amount of communication rounds determines the communication overhead.
Together with the maximum utility or privacy loss that can be tolerated by the target application, 
it would help to find the sweet spot point on this trade-off curve.

\vspace{-4mm}
\section{Deploying Blinder on Mobile and Edge Devices}
\subsection{Blinder's Mobile Application}
\label{subsubsec:android}
We develop an Android app and deploy it on smartphones 
to measure Blinder's anonymization latency and power consumption in the real world. 
Blinder's VAE is trained via the proposed personalized federated learning algorithm 
using PyTorch~\cite{NEURIPS2019_9015}.
The pre-trained model is quantized and serialized 
such that it becomes compatible with the PyTorch Mobile framework. 
The quantized encoder and decoder have the same size of $6 MB$.
We select $3$ Android smartphones %
with diverse computing powers. They are listed in Table~\ref{tab:smartphone}.
Our Android app pulls sensor readings from the smartphone's onboard accelerometer and gyroscope.
The sampling rate of both the accelerometer and gyroscope is set to $50 Hz$.
Thus, to achieve real-time data anonymization, 
Blinder must complete the anonymization of a data segment 
before the next data segment is ready, i.e., within a $200$ ms interval.
We use the pre-processing techniques described in Section~\ref{subsec:datasets}.
Since the public attribute needs to be appended to the learned latent representation, 
a pre-trained inference model is utilized to predict the user's public attribute 
from the original sensor data segment.
In our experiments, we reuse the desired inference model 
described in Section~\ref{subsec:eval_utility} to predict the public attribute.
The overhead of running this inference model is measured 
and lumped with the execution time of other pre-processing steps. 
We refer to this total time as preparation time. %
The private attribute passed to the decoder is chosen randomly for stochastic anonymization.

We use PowerTutor~\cite{zhang2010accurate}, 
an open-source app that takes accurate app-level CPU power consumption measurements, 
to estimate the battery drain due to the execution of Blinder. %
We evaluate the performance of Blinder when trained for weight anonymization 
on MobiAct. %
The total anonymization latency is reported 
in Table~\ref{tab:smartphone}. 
The latency measurement is averaged over anonymizing $1000$ sensor data segments %
and the power consumption is averaged over $20+$ minutes of its execution.
We find that Blinder's total anonymization latency is $8.81$ ms on average 
among the 3 Android smartphones,
which is $\sim{22}\times$ faster than our real-time anonymization budget.
Furthermore, the smartphone's CPU consumes battery at around $6.38$ J/min when running Blinder, 
which is around $4\times$ its idle power consumption. 
To put it in context, Blinder consumes around $1.5-2\times$ more power than the Google Maps app.
Note that we use a $50 Hz$ sampling rate in our experiment, 
which is $10\times$ faster than the default sampling rate used in Android. 
Thus, Blinder's power consumption can be further reduced by adjusting the sampling rate 
according to the data need of the target application.
We also remark that since the discriminator is just used to train Blinder
and is not loaded when anonymizing sensor data, 
protecting multiple private attributes is not expected to increase Blinder's 
power draw or computational overhead.

\subsection{Anonymization on IoT Edge Devices}
Blinder is also deployed on NVIDIA Jetson Nano, which is a representative IoT edge device. %
Jetson is a power-efficient computing platform, 
ideal for running applications that could benefit from GPU acceleration.
Our Jetson Nano model has $2$ GB of RAM that is shared by a quad-core CPU and a 128-core GPU. 
It runs Linux4Tegra (L4T) OS and NVIDIA Jetpack 4.4 SDK.
We installed the PyTorch for Jetson library to allow the pre-trained Blinder model 
to be effortlessly deployed for real-time anonymization.
Since Jetson does not come with IMU sensors, 
we simulate sensor data generation using the APScheduler library~\cite{apscheduler-git}.
We perform two sets of experiments: utilizing only the CPU and enabling Jetson's GPU acceleration.
Other aspects of these experiments are identical to the ones described in Section~\ref{subsubsec:android}.
As Table~\ref{tab:smartphone} shows, when utilizing the CPU only, 
Jetson can complete the anonymization task in $34.45$ ms on average, 
which is much less than our $200$ ms time budget. 
By enabling the GPU acceleration capability, 
the total anonymization latency decreases by over $40\%$ to $14.76$ ms on average. 
It should be noted that the total anonymization latency on Jetson is higher than on smartphones. 
We believe this is because the OS, required services, and Blinder are using $2$ GB of shared RAM.
As a result, running Blinder on Jetson requires frequent usage of the swap memory, 
which runs off a microSD card and has slower I/O performance than the eMMC/UFS-based storage in smartphones.
We do not report Blinder's power draw because %
Jetson has a reliable power supply (via a USB connector) 
and is not powered by a battery. 

\begin{table}[tb]
    \centering
    \resizebox{\columnwidth}{!}{
    \begin{tabular}{l  l  c  c  c  c  c} 
     \textbf{Device Name} & \textbf{Processor} & Total (ms) & Preparation (ms) & Encoder (ms) & Decoder (ms)\\
     \toprule
     OnePlus 6  & Snapdragon 845 & 10.84 & 7.55 & 1.83 & 1.47 \\
     \midrule
     Nokia 6.1 Plus  & Snapdragon 636 & 7.05 & 3.90 & 1.69 & 1.46 \\
     \midrule
     Samsung Note 4 & Exynos 5433 & 8.55 & 4.10 & 2.38 & 2.06 \\
     \midrule
     NVIDIA Jetson Nano & Arm Cortex-A57 & 34.45 & 22.47 & 6.16 & 5.82 \\
     \midrule
     NVIDIA Jetson Nano & Maxwell GPU & 14.76 & 6.57 & 3.68 & 4.51 \\
     \bottomrule
    \end{tabular}%
    }
    \vspace{2mm}
    \caption{Anonymization latency of Blinder in the weight anonymization task. %
    }\label{tab:smartphone}%
    \vspace{-9mm}
\end{table}

\section{Limitations and Discussion}
The premise of this work is that there are a number of 
well-defined private attributes that users aim to conceal from a passive adversary 
(described in Section~\ref{sec:problem_def})
that has access to their sensor data, 
for example to make inferences that are valuable to them.
One limitation of our work is that 
Blinder cannot be trained when private attributes are not well-defined 
or users deem everything that can be inferred from their sensor data private, 
except for a single public attribute.
Additionally, several privacy risks have been identified in distributed and federated learning 
that largely stem from sharing model parameters or gradients with a central server or other clients.
For example, a passive adversary can perform the model inversion attack 
to recover the private attribute under certain conditions, 
and a more powerful malicious adversary can reduce Blinder's privacy-preserving 
capability through a model poisoning attack.
We do not study these attacks and other adversary models in this work.
In future work, we plan to investigate the potential privacy loss 
due to sharing gradients with the server in the personalized federated learning setting.

\section{Conclusion}
This paper proposes a novel data anonymization model that can be deployed on mobile and IoT edge devices. 
Built on top of a variational autoencoder, Blinder learns to extract latent representations 
that can be conveniently modified to obscure users' private information. 
We apply meta-learning to federated learning to enable a number of clients 
to collaboratively train Blinder, thereby eliminating the need 
for sending users' raw sensor data and attributes
to an HBC adversary and providing end-to-end privacy protection.
Blinder employs a public attribute rebalancing technique and a generative model 
to tackle the problems related to imbalanced, non-i.i.d. data in the federated learning setting.
Our extensive evaluation suggests that Blinder has strong anonymization capability, 
can effectively reduce the accuracy of intrusive inferences to nearly the same level as random guessing,
generalizes well to heterogeneous data and new users,
and can be extended to obscure multiple user-defined private attributes simultaneously.
Furthermore, real-world deployment of Blinder on smartphones and a representative edge device
confirms that it is suitable for performing real-time anonymization 
of timeseries generated by embedded sensors.

\balance

\begin{acks}
This research was supported by funding from the Natural Sciences and Engineering Research Council of Canada (RGPIN-2019-04349).
\end{acks}

\bibliographystyle{ACM-Reference-Format}
\bibliography{paper}

\end{document}